\documentclass[11pt]{article}

\usepackage[]{acl}

\usepackage{times}
\usepackage{latexsym}
\usepackage{amsmath} 
\usepackage{multirow} 
\usepackage{algorithm}
\usepackage{algorithmic}
\usepackage{amsmath, amssymb}      
\usepackage{amsfonts}

\usepackage[T1]{fontenc}

\usepackage[utf8]{inputenc}

\usepackage{microtype}

\usepackage{inconsolata}

\usepackage{graphicx}

\title{Robust Uncertainty Quantification for Self-Evolving Large Language Models via  Continual Domain Pretraining}

\newcommand{\squishlist}{
  \begin{list}{$\bullet$}{
    \setlength{\itemsep}{0pt}
    \setlength{\parsep}{3pt}
    \setlength{\topsep}{3pt}
    \setlength{\partopsep}{0pt}
    \setlength{\leftmargin}{1.5em}
    \setlength{\labelwidth}{1em}
    \setlength{\labelsep}{0.5em}
  }
}

\newcommand{\squishlisttwo}{
  \begin{list}{$\bullet$}{
    \setlength{\itemsep}{0pt}
    \setlength{\parsep}{0pt}
    \setlength{\topsep}{0pt}
    \setlength{\partopsep}{0pt}
    \setlength{\leftmargin}{2em}
    \setlength{\labelwidth}{1.5em}
    \setlength{\labelsep}{0.5em}
  }
}

\newcommand{\squishend}{
  \end{list}
}
\usepackage{float}

\author{
 \textbf{Xiaofan Zhou},
 \textbf{Lu Cheng},
\\
 \textsuperscript{}University of Illinois at Chicago,
\\
 \small{
   \textbf{Correspondence:} \href{xzhou77@uic.edu}{xzhou77@uic.edu}
 }
}

\begin{document}
\maketitle
\begin{abstract}
Continual Learning (CL) is essential for enabling self-evolving large language models (LLMs) to adapt and remain effective amid rapid knowledge growth. Yet, despite its importance, little attention has been given to establishing statistical reliability guarantees for LLMs under CL, particularly in the setting of continual domain pretraining (CDP). Conformal Prediction (CP) has shown promise in offering correctness guarantees for LLMs, but it faces major challenges in CDP: testing data often stems from unknown or shifting domain distributions, under which CP may no longer provide valid guarantees. Moreover, when high coverage is required, CP can yield excessively large prediction sets for unanswerable queries, reducing informativeness.
To address these challenges, we introduce an adaptive rejection and non-exchangeable CP framework. Our method first estimates the distribution of questions across domains in the test set using transformer-based clustering, then reweights or resamples the calibration data accordingly. Building on this, adaptive rejection CP allows the LLM to selectively abstain from answering when its confidence or competence shifts significantly. Extensive experiments demonstrate that our framework enhances both the effectiveness and reliability of CP under CDP scenarios. Our code is available at: https://anonymous.4open.science/r/CPCL-8C12/
\end{abstract}
\section{Introduction}
LLMs have demonstrated impressive performance across diverse domains, from personal assistants \cite{achiam2023gpt} and robotics \cite{wang2025large} to biology \cite{jung2024llm} and scientific discovery \cite{rane2023contribution}. Yet, in the face of rapid knowledge growth, new and private information emerges at a pace that quickly renders their static, pre-trained knowledge outdated. This dynamic landscape underscores the need for self-evolving LLMs that can continuously adapt to novel environments. CL has therefore become a critical capability, allowing LLMs to incorporate new knowledge while preserving previously acquired skills \cite{shi2024continual, gururangan2021demix, li2023cfgpt, lu-etal-2025-controlled}. 

As research on continual learning (CL) for LLMs progresses—particularly in high-stakes domains—the need to robustly quantify uncertainty during the self-evolving process becomes paramount. A central paradigm in uncertainty quantification (UQ) is ensuring that model outputs cover the ground-truth answer with high probability. Conformal Prediction (CP) \cite{zhou2025conformal, shafer2008tutorial, angelopoulos2021gentle} offers such a framework: a model-agnostic method that provides formal statistical guarantees that the correct answer is included with a user-specified probability. In recent years, CP has been successfully applied across diverse LLM tasks, including multiple-choice QA \cite{kumar2023conformal, ye2024benchmarking}, open-ended QA \cite{su-etal-2024-api, li-etal-2024-traq}, long-form answering \cite{mohri2024language, rubin2025conformal}, and even LLM-as-a-judge evaluations \cite{sheng2025analyzinguncertaintyllmasajudgeinterval}, substantially enhancing reliability.

\begin{figure}
    \centering
    \includegraphics[width=1\linewidth]{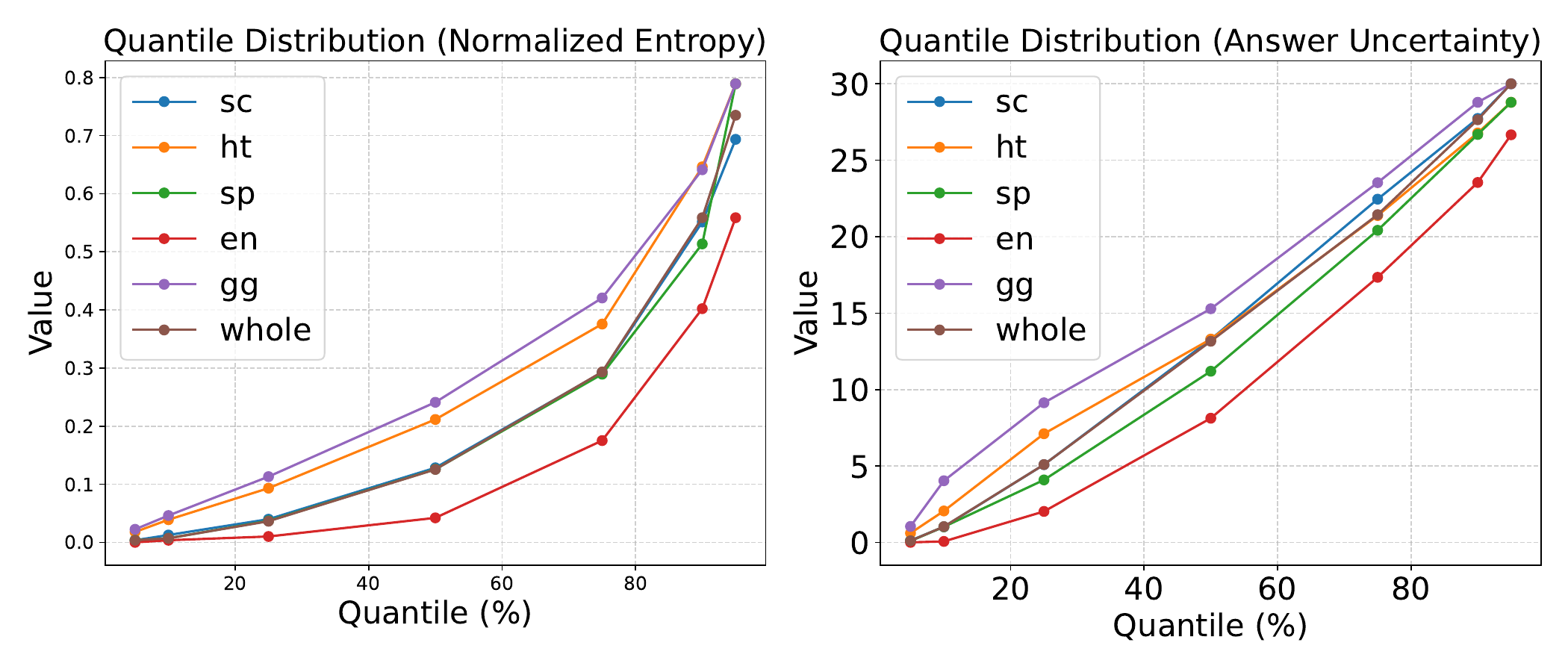}
    \caption{Different quantiles across domains (i.e., ``sc'', ``ht'', ``sp'', ``en'', ``gg'') in TriviaQA for NE (left), indicating uncertainty about whether the LLM can answer, and the answer uncertainty (right), measuring the likelihood that the LLM will produce the ground-truth answer. $x$-axis is quantile and $y$-axis is quantile value for the uncertainty measurement. ``whole'' represents average quantile. We see that quantile values vary among different domains. Results are from Llama-3.1-8B-Instruct.}
    \label{fig:p12}
\end{figure}
In this work, we study open-ended QA under continual domain pretraining (CDP), a CL setting where LLMs are continually fine-tuned on knowledge from diverse domains. Applying CP to CDP introduces two major challenges: (1) \textit{Violation of Exchangeability.} CP’s coverage guarantees rely on exchangeability (e.g., I.I.D. data). In CDP, selecting fine-tuning data across domains can break this assumption, leading to uneven coverage across domains (Figure \ref{fig:p12}). To address this, we propose a non-exchangeable CP framework that leverages a buffer dataset and calibrates using the covariate distribution of the test data. (2)
\textit{Uninformative Prediction Sets.} CP may produce excessively large sets when LLMs face uncertainty in new domains, making predictions uninformative. While adding “Can’t answer” labels can help \cite{li-etal-2024-traq}, naive thresholding fails under catastrophic forgetting \cite{toneva2018empirical}. We develop an adaptive CP method based on label-conditional CP \cite{lofstrom2015bias}, which dynamically adjusts to shifting model capabilities and achieves optimal efficiency by producing the smallest prediction sets.

Overall, our main contributions are:
\squishlist
\item To the best of our knowledge, this work is the first to investigate CP in the context of CDP for LLMs. We introduce a model-agnostic uncertainty quantification framework tailored to CDP, enabling reliable adaptation of LLMs across evolving domains.

\item We address two key challenges of robust UQ for CDP via CP: non-exchangeability and adaptive rejection under catastrophic forgetting. In addition, we provide a theoretical analysis for guaranteeing better performance of our approach (Appendix \ref{theory}).

\item Extensive experiments demonstrate that our approach --AR-NECP--effectively mitigates the distribution shift between calibration and testing data, enabling LLMs to reject unanswerable questions while maintaining reliable correctness coverage.
\squishend

\section{Related work}
\subsection{CDP in LLM}
CDP fine-tunes LLMs sequentially across domains \cite{shi2024continual}. \citet{gururangan-etal-2020-dont} show that domain-adaptive and task-adaptive pretraining (DAPT, TAPT) consistently enhance downstream performance. To mitigate domain interference, \citet{gururangan2021demix} proposes DEMix layers with modular experts that disentangle representations. \citet{qin2022elle} present ELLE, which expands model capacity using domain prompts for efficient lifelong adaptation, while recyclable tuning \cite{qin2023recyclable} reuses prior weights via distillation to speed convergence and retain knowledge. \citet{han2020econet} introduces ECONET for continual pretraining on temporal and event reasoning. \citet{lu-etal-2025-controlled} propose a novel Lora adaptor for dealing with catastrophic forgetting. Domain-specific variants such as PMC-LLaMA \cite{wu2024pmc}, SaulLM-7B \cite{colombo2024saullm}, Lawyer LLaMA \cite{huang2023lawyer}, EcomGPT-CT \cite{ma2023ecomgpt}, CFGPT \cite{li2023cfgpt}, and AF Adapter \cite{yan2023af} further realize vertical continuity, tailoring the pretraining–fine-tuning pipeline to specialized domains while balancing transfer and forgetting.
\subsection{CP for LLM}
CP enables large language models (LLMs) to produce reliable prediction sets or calibrated outputs \cite{zhou2025conformal, campos2024conformal}. Existing CP methods for LLMs fall into three categories: (1) sampling-based CP, which samples multiple responses and selects those exceeding a calibrated uncertainty threshold \cite{su-etal-2024-api, li-etal-2024-traq, gui2024conformal}, though its achievable coverage is limited by model accuracy; (2) adaptive-sampling CP, which continues sampling until reaching the desired coverage \cite{quach2023conformal} but at high computational cost \cite{su-etal-2024-api}; and (3) abstractive CP, which replaces uncertain content with generic statements in generative tasks \cite{mohri2024language, rubin2025conformal}. Other applications include LLM-as-a judge \cite{sheng2025analyzinguncertaintyllmasajudgeinterval}, classification, Multi-choice QA \cite{kumar2023conformal, ye2024benchmarking}. However, all these methods assume exchangeability between calibration and test data, which often breaks under domain shifts or CL. As the first CP method for CDP, our method provides approximate finite-sample or \textit{global} asymptotic coverage guarantees across domains, reduces computation by calibrating once per dataset instead of per instance, and extends CP to LLMs' CDP---achieving scalable, domain-aware, and more reliable calibration under domain shifts. The most related work is \cite{gaomodel} that proposes a nearest-neighbor calibration achieving \textit{local} asymptotic coverage under traditional CL settings.
\section{Preliminary}
\subsection{Background in CP}

CP is a model-agnostic framework that provides distribution-free uncertainty quantification with guaranteed coverage \cite{shafer2008tutorial, angelopoulos2021gentle}. Given a data set $\{(x_i, y_i)\}_{i=1}^n$, it first divides it into training data $D_{train}\{(x_i, y_i)\}_{i=1}^{n_{train}}$, $D_{cal}\{(x_i, y_i)\}_{i=1}^{n_{cal}}$ and $D_{test}\{(x_i, y_i)\}_{i=1}^{n_{test}}$, where $x_i$ is the question, $y_i$ is the corresponding answer to $x_i$, $n_{*}$ is the number of samples in the corresponding dataset, $n$ = $n_{train}+n_{cal}+n_{test}$.

Let $S(f, (x,y))$ denote a nonconformity score that measures how unusual label $y$ is for input $x$ under model $f$. For example, in classification, it is usually calculated by:
\begin{align}
S(f, (x,y)) = 1-f_y(x),
\end{align}
where $f_y(x)$ is probability of $f$ predict $x$ as $y$. 

Using the calibration set $\{(x_i, y_i)\}_{i=1}^{n_{cal}}$, it compute non-conformity scores as:
\begin{align}
s_i = S(f, (x_i, y_i)), \quad i=1,\dots,n_{cal}.
\end{align}
Based on the above equation, it calculates a threshold quantile for including labels by:
\begin{align}
\label{text_quantile}
q_{\alpha}
= Quantile\!\left( \{s_i\}_{i=1}^{n_{cal}};\, \frac{(n_{cal}+1)(1-\alpha)}{n_{cal}} \right).
\end{align}
For a new test input $x_{n+1}$ and candidate labels $y\in Y$, we compute the non-conformity score for $y$ as $s_{n+1}(y) = S(f, (x_{n+1}, y))$. 
Then the prediction set is defined as:
\begin{align}
C_\alpha(x_{n+1}) = \{ y : s_{n+1}(y) < q_{\alpha} \},
\end{align}
which guarantees finite-sample coverage:
\begin{align}
\mathbb{P}\big( y_{n+1} \in C_\alpha(x_{n+1}) \big) \geq 1 - \alpha.
\end{align}
Besides the coverage guarantee, CP also needs to ensure a small prediction set, i.e., high prediction efficiency, to be informative
\subsection{Sampling-based CP in LLMs}
One common strategy of applying CP in LLMs is to use sampling-based methods, especially for open-ended QA tasks \cite{su-etal-2024-api, li-etal-2024-traq}. Given an LLM, one would only need two datasets, calibration data $D_{cal}\{(X_i, Y_i)\}_{i=1}^{n_{cal}}$ and $D_{test}\{(X_i, Y_i)\}_{i=1}^{n_{test}}$.

It involves two stages, the calibration stage and the testing stage. In the calibration stage, for each data point $(X_i, Y_i)$, it first asks the LLM to generate the answer for $M$ times, and uses a similarity-based method, e.g., Rouge-L score, cosine similarity, to cluster the answers to get a clustered answer set:
\begin{align}
\label{cluster}
C_i=\{c_{ij} |j\in [1, N]\},
\end{align}
where $c_ij$ is the representative answer in the cluster $j$ for question $i$ and $N$ is the number of clusters. Then it uses the number of answers in each cluster as the uncertainty score, as the clustering may not be perfect, and each question may have various answers. It then calculates the non-conformity score as:
\begin{align}
&S(f,(x,y)) = \\ \nonumber
&\max\limits_{j \in [1,M],\, c_j=y} \Bigl(1-\tfrac{\text{freq}_j}{M}\Bigr),
\quad \text{if such $j$ exists},\\ \nonumber
& S(f,(x,y)) = 0, 
\quad \text{otherwise}.
\end{align}
 The quantile is calculated by Eq. \ref{text_quantile}. In the inference stage, the prediction set is built by:
\begin{align}
C_\alpha(x_{n+1}) = \{ c_j : 1-\frac{\text{freq}_{j}}{M} < q_{\alpha}\}.
\end{align}
The limitation of this method is that the target coverage is upper-bounded by the model's ability.

\section{Proposed Approach}

\subsection{Problem Formulation}
We consider $K$ QA datasets ${D^k}, k\in[1,...,K]$, each from different domains, and split them into ${D^k_{train}, D^k_{buffer}}$, where each dataset $D^k_{train}$ is used to finetune a corresponding LLM $\phi_k$, $D^k_{buffer}$ is used to calibrate CP method. At each training step, $\phi_k$ has only limited access to data from previous tasks, which is a common setting in CDP. For illustration simplicity, we describe our proposed CP method on the final finetuned model $\phi_K$. At this stage, a union dataset $D_{\text{test}}$ containing all domains from previous steps is provided, without domain labels, and an unknown domain distribution. \textbf{Our goal is to enable LLM $\phi_K$ to generate prediction sets that marginally guarantee to cover the right answer with 1-$\alpha$ probability.}
\\
\textbf{Exchangeable Assumption within Domains.}
Our problem setting assumes that data within each domain is exchangeable following \cite{li-etal-2024-traq}, i.e., data from calibration and testing sets in the same domain are drawn independently and identically from the same data distribution $D$. 
\\
\textbf{Overview.} Our framework follows protocols in sampling-based CP and comprises two key components: the \textit{Adaptive Rejection CP} and the \textit{Non-Exchangeable CP} modules. 
The Adaptive Rejection CP introduces a dynamic rejection mechanism that adaptively determines the threshold for abstaining from answering while maintaining the ground truth within the prediction set, thereby ensuring the desired coverage guarantee. 
The Non-Exchangeable CP component first estimates the domain composition of the testing data by analyzing grouping statistics between the calibration and testing sets. It then either resamples or reweights the calibration samples based on the clustering results from both sets, so that the distribution of nonconformity scores in the calibration data closely aligns with that of the testing data. 
This adjustment statistically restores exchangeability between the two distributions, enabling reliable coverage under domain shift. 
The overall procedure of the proposed algorithm is summarized in Algorithm~\ref{alg:arcrcp}. A theoretical analysis of our method for guaranteeing better CP performance can be seen in Appendix \ref{theory}.

\subsection{Adaptive Rejection CP}
\label{arcp}
A simple abstention baseline assigns a `can’t answer' label using a lightweight MLP’s predicted probability \cite{li-etal-2024-traq}, but it lacks a coverage-aware, principled rejection mechanism. We enable refusal via a \textbf{label-conditional CP }scheme calibrated to maintain coverage \(1-\alpha\) on answerable questions; items labeled `can’t answer’’ are then rejected. Concretely, we fix the error rate on answerable instances at ($\alpha$)—matching the global CP level—to yield well-calibrated prediction sets. To mitigate catastrophic forgetting across domains, we incorporate the estimated \textit{answerability rate} from calibration data to adaptively set the rejection threshold. We detail the procedure next.
\\
\textbf{Rejection Mechanism}
To begin with, each sample $x_i \in D_{cal}$ in the dataset is given an unanswerable score $s_i$. In our project, we propose to use normalized entropy (NE) as the unanswerable score, which is a statistical metric to predict the LLM's ability to answer a question. It is calculated as:
\begin{align}
NE(x_i) &= - \frac{1}{\log K} \sum_{j=1}^{K} |c_j| \log |c_j|,
\end{align}
where $|c_j|$ is the number of answers in cluster $c_j$, and $K$ is the total number of clusters that $x_i$ has. $NE \in [0,1]$, where higher $NE$ indicates lower confidence to predict the question. We further formulate the probability that the LLM cannot ($p^0$)/can ($p^1$) answer the question as:
\begin{align}
p^0(x_i) = NE(x_i);
p^1(x_i) = 1 - NE(x_i).
\end{align}

Setting the error rate of unanswerable questions as $\alpha_0$ and the error rate of answerable questions as $\alpha_1$, we can get the overall coverage as:
\begin{align}
\alpha = 
 (1 - r_{correct}) \alpha_0 + r_{correct} (\alpha_1 + \alpha-\alpha\alpha_1),
 \label{eq:correct}
\end{align}
where $r_{correct}$ is the ratio of answerable questions in the calibration dataset. Under the exchangeability assumption within domains, Eq. \ref{eq:correct} remains valid for the test set. We consider a question as answerable if any sampled question's answer regex matches the ground truth. Thus, we can get the relation between $\alpha_0$ and $\alpha_1$ as:
\begin{align}
\label{a01}
 \alpha_1 = \frac{(1 - r_{correct})(\alpha - \alpha_0)}{r_{correct}(1-\alpha)}.
\end{align}

After this, we denote the answerable question as $x^1$ and the unanswerable question as $x^0$, with $n^0_{cal}$ as the number of questions that cannot be answered and $n^1_{cal}$ as the number of questions that can be answered in the calibration data. We calculate the quantile for adding the `can't answer' label as:
\begin{align}
\label{q0}
&q^0_{\alpha_0} = \\ \nonumber
& \operatorname{Quantile}\!\left( \{p^0(x^0_i)\}_{i=1}^{n^0_{cal}};\, \frac{(n^0_{cal}+1)(1-\alpha_0)}{n^0_{cal}} \right),
\end{align}
and the quantile for rejecting to answer as:
\begin{align}
\label{q1}
&q^1_{\alpha_1} = \\ \nonumber
& \operatorname{Quantile}\!\left( \{p^1(x^1_i)\}_{i=1}^{n^1_{cal}};\, \frac{(n^1_{cal}+1)(1-\alpha_1)}{n^1_{cal}} \right).
\end{align}

In the testing stage, LLM refuses to answer the question when $p^0(x_i) < q^0_{\alpha_0}$ and $p^1(x_i) > q^1_{\alpha_1}$, where the NE score shows that the LLM has a low probability of answering the question but a high probability of being unanswerable. We add the `can't answer' label to the prediction set when $p^0(x_i) < q^0_{\alpha_0}$ and $p^1(x_i) < q^1_{\alpha_1}$ to ensure a guaranteed coverage. For the remaining cases, where the LLM has a low probability of being unanswerable, we use the standard CP method.
\\
\textbf{Prediction Set for Answerable Questions}
After labeling a portion of questions as unanswerable, some truly answerable questions may be mistakenly classified as unanswerable. This misclassification introduces a distribution bias between the set of all answerable questions and the subset predicted as answerable.
To mitigate this bias, we recompute the quantile to ensure that the prediction set properly includes the retained answerable questions, as follows:
\begin{align}
\label{text_quantile_1}
\hat{q}_{\alpha}^{text}
= \operatorname{Quantile}\!\left( \{\hat{s}_i\}_{i=1}^{\hat{n}_{cal}};\, \frac{(\hat{n}_{cal}+1)(1-\alpha)}{\hat{n}_{cal}} \right),
\end{align}
where the frequency-based non-conformity score $\hat{s}_i$ corresponds to the answerable questions with $p^1(x_i) < q^1_{\alpha_1}$, and $\hat{n}_{cal}$ is the number of such questions. For answerable questions, clusters with higher cluster scores than $\hat{q}_{\alpha}^{text}$ are included in the prediction set $c_i$, if not rejected.
Due to the concentration issue caused by using frequency alone \cite{su-etal-2024-api}, we propose to set the cluster scores as frequency minus NE.
\\
\textbf{Grid Search for Best Prediction Efficiency}
With the above quantiles, we use a grid search to achieve the best efficiency (i.e., smallest prediction set), which is especially helpful when the answerable rate is high or the coverage requirement is large, making it hard to find unanswerable questions. Specifically, we choose the pair of $(\alpha_0, \alpha_1)$ according to Eq.~\ref{a01}. This can be formulated as:
\begin{align}
\alpha_0, \alpha_1 = \arg\min_{\alpha_0, \alpha_1} E(|C_\alpha(x_i)| \mid x_i \in D_{cal}),
\end{align}
where $|C_\alpha(x_i)|$ is the number of answers in the prediction set for $x_i$.
\subsection{Non-Exchangeable CP Across Domains}
\label{resample}
In CDP, data from different domains can be non-exchangeable. To achieve reliable coverage across different domains, after each domain’s pretraining, we draw a small subset from $D^k$ as $D^k_{\text{buffer}}$. We further split it into two parts, $D^k_{cluster}$ and $D^k_{\text{cal}}$, with sizes $n^k_{cluster}$ and $n^k_{\text{cal}}$, respectively. $D^k_{cluster}$ will be used for constructing a cluster centroid for each domain, and $D^k_{\text{cal}}$ will be used to calibrate for CP. Given a joint test question distribution $D_{\text{test}}$, the domain information is unknown in CDP. Moreover, the exchangeability assumption no longer holds in this setting as testing data and calibration data may have different domain distributions. To address these challenges, we first approximate the domain distribution in the test data and then resample the calibration data accordingly. Our method mitigates the domain distribution shift under the assumption that within each domain, data is exchangeable.
\\
\textbf{Domain Cluster Centroids via Transformer Encoders.}
\label{domain}
We use a Transformer encoder $T(\cdot)$ to obtain semantic embeddings for all questions, representing each domain in a shared embedding space. 
For domain $k$ with $n_{\text{cluster}}^k$ questions $\{x_i^k\}_{i=1}^{n_{\text{cluster}}^k}$, each question is encoded as
\begin{align}
\mathbf{h}_i^k = T(x_i^k) \in \mathbb{R}^d.
\end{align}
The domain centroid is then the mean embedding:
\begin{align}
\label{centroid}
\mathbf{c}_k = \frac{1}{n_{\text{cluster}}^k}\sum_{i=1}^{n_{\text{cluster}}^k}\mathbf{h}_i^k,
\end{align}
which is further $\ell_2$-normalized for comparability:
\begin{align}
\tilde{\mathbf{c}}_k = \frac{\mathbf{c}_k}{\|\mathbf{c}_k\|_2}.
\end{align}
A new question $x_i$ is assigned to the domain with the highest cosine similarity:
\begin{align}
c(x_i) = \arg\max_{k \le K}(\tilde{\mathbf{c}}_k^\top \mathbf{h}_i).
\label{eq:assignclu}
\end{align}
\\
\textbf{Testing data domain distribution estimation from the Buffer.}
\label{samplling}
After obtaining the cluster centroid for each domain, we group both calibration and testing data according to their nearest domain centroids. Under the I.I.D. assumption within each domain, the statistics of this clustering on the calibration and testing data can be used to estimate the proportion of domains in the testing set.

In the calibration data, after clustering, we obtain each domain’s cluster assignment result. We denote $P_{ij}$, where $i,j \in \{1, \dots, K\}$, as the ratio of questions from domain $i$ that are clustered into domain $j$. Each domain in the test set will follow a similar transition pattern:
\begin{align}
    \tilde{n}_{\text{test}}^j = \sum_{i=1}^{K} \hat{n}_{\text{test}}^i P_{ij} ,
\end{align}
where $\tilde{n}_{\text{test}}^j$ is the number of test samples clustered into domain $j$. 
Let $P = [P_{ij}]_{i,j=1}^{K}$, the number of questions in the testing data clustered into domain $k$ as $\tilde{\mathbf{n}}_{\text{test}} = [\tilde{n}_{\text{test}}^{1}, \dots, \tilde{n}_{\text{test}}^{K}]^{\top}$, and estimate true number of questions in each domain $k$ in $\hat{\mathbf{n}}_{\text{test}} = [\hat{n}_{\text{test}}^{1}, \dots, \hat{n}_{\text{test}}^{K}]^{\top}$.  
The system can then be written compactly as:
\begin{align}
P^{\top}\hat{\mathbf{n}}_{\text{test}} = \tilde{\mathbf{n}}_{\text{test}}.
\end{align}
As \( P^{\top} \) is a diagonally dominant matrix, it is invertible \cite{horn2012matrix}. We can then estimate the true domain counts by:
\begin{align}
\hat{\mathbf{n}}_{\text{test}} = (P^{\top})^{-1}\tilde{\mathbf{n}}_{\text{test}}.
\label{eq:invert}
\end{align}
Since the transformer-based clustering method demonstrates strong separability performance \cite{wu2024transformer}, Eq. \ref{eq:invert} holds in practice.
\\
\textbf{Resampling CP}
After estimating the number of questions in each domain of the testing data, we resample the calibration data from each domain to form $\hat{D}^k_{\text{cal}}$. The union $\hat{D}^{\text{union}}_{\text{cal}}$ thus approximates the domain distribution of the testing dataset. 
\\
\textbf{Reweighing CP}
Another approach for alleviating distribution shift in CP is Weighted CP \cite{tibshirani2019conformal}, which reweighs calibration samples based on the estimated domain proportions. Let $\{s_i, w_i\}_{i=1}^{n_{\text{cal}}}$ denote the nonconformity scores and their associated weights, where for each calibration sample $x_i \in D_{\text{cal}}^k$, the weight is defined as
\begin{align}
w_i = \frac{\hat{n}_{\text{test}}^k}{\sum_{k=1}^{K} n_{\text{test}}^k}.
\end{align}
The weighted quantile threshold $\hat{q}_\alpha$ is then
\begin{align}
\hat{q}_\alpha = \inf\left\{ q \in \mathbb{R} : \frac{\sum_{i=1}^{n_{\text{cal}}} w_i \mathbf{1}(s_i \le q)}{\sum_{i=1}^{n_{\text{cal}}} w_i} \ge 1 - \alpha \right\},
\end{align}
where $\mathbf{1}(\cdot)$ denotes the indicator function. In the Adaptive Rejection CP framework, all thresholds $q_{\alpha_0}^0$, $q_{\alpha_1}^1$, and $\hat{q}_{\alpha}^{\text{text}}$ are recalculated accordingly based on the weighted calibration scores. Compared with resampling-based strategies, Weighted CP provides a deterministic adjustment and generally yields more stable coverage when certain domains in the test set are severely underrepresented, as resampling may fail to include sufficient representative samples from minority domains.


\begin{algorithm}[t]
\caption{Adaptive Rejection and Non-Exchangeable CP (AR-NECP)}
\label{alg:arcrcp}
\begin{algorithmic}[1]
\REQUIRE Split $D^k_{buffer}$ into $\{D^k_{cluster}, D^k_{\text{cal}}\}_{k=1}^{K}$, $x_{\text{test}}$ in $D_{\text{test}}$, LLM $\phi_K$, significance $\alpha$, transformer model $T$
\ENSURE Prediction sets $C_{\alpha}(x)$

\STATE \textbf{Step 1: Domain Centroids}
\FOR{$k = 1, \dots, K$}
    \STATE Get $c_k$ from $D_{cluster}^k$ using Eq. \ref{centroid}.
\ENDFOR

\STATE Assign $x \in D_{\text{cal}} \cup D_{\text{test}}$ using Eq. \ref{eq:assignclu}.

\STATE \textbf{Step 2: Non-exchangeable CP}
\STATE Compute $P_{ij}$ and $\tilde{\mathbf{n}}_{\text{test}}$ from clustering
\STATE Estimate $\hat{\mathbf{n}}_{\text{test}}=(P^{\top})^{-1}\tilde{\mathbf{n}}_{\text{test}}$
\STATE Resample or reweight $\hat{D}^k_{\text{cal}}$ according to $\hat{\mathbf{n}}_{\text{test}}$

\STATE \textbf{Step 3: Adaptive Rejection CP Calibration}
\STATE Calculate $q_{\alpha_0}^{0}$, $q_{\alpha_1}^{1}$, $\hat{q}_{\alpha}^{text}$ for each domain clustered dataset $\hat{D}^{union}_{cal}$ by method in Section \ref{arcp}.

\STATE \textbf{Step 4: Inference}
\FOR{$x_i \in D_{\text{test}}$}
    \STATE Build $C_{\alpha}(x_i)$ with $q_{\alpha_0}^{0}$, $q_{\alpha_1}^{1}$, $\hat{q}_{\alpha}^{text}$ by algorithm described in Section \ref{arcp}.
\ENDFOR 
\STATE \RETURN $C_{\alpha}(x)$
\end{algorithmic}
\end{algorithm}

\section{Experiment} 
For a comprehensive evaluation, we focus on the following research questions:
\squishlist
\item \textbf{RQ. 1}. How effective is our AR-NECP method in addressing the distribution shift in CDP?
\item \textbf{RQ. 2}. How does the adaptive rejection CP improve the prediction efficiency of CP (especially on the unanswerable questions)? 
\item \textbf{RQ. 3}. How do different UQ methods for LLM answer rejection influence CP's prediction efficiency?
\item \textbf{RQ. 4}. How well does $\hat{n}^k_{\text{test}}$ (estimated number of test questions per domain) work in practice? 
\squishend
\begin{figure}
    \centering
    \includegraphics[width=1\linewidth]{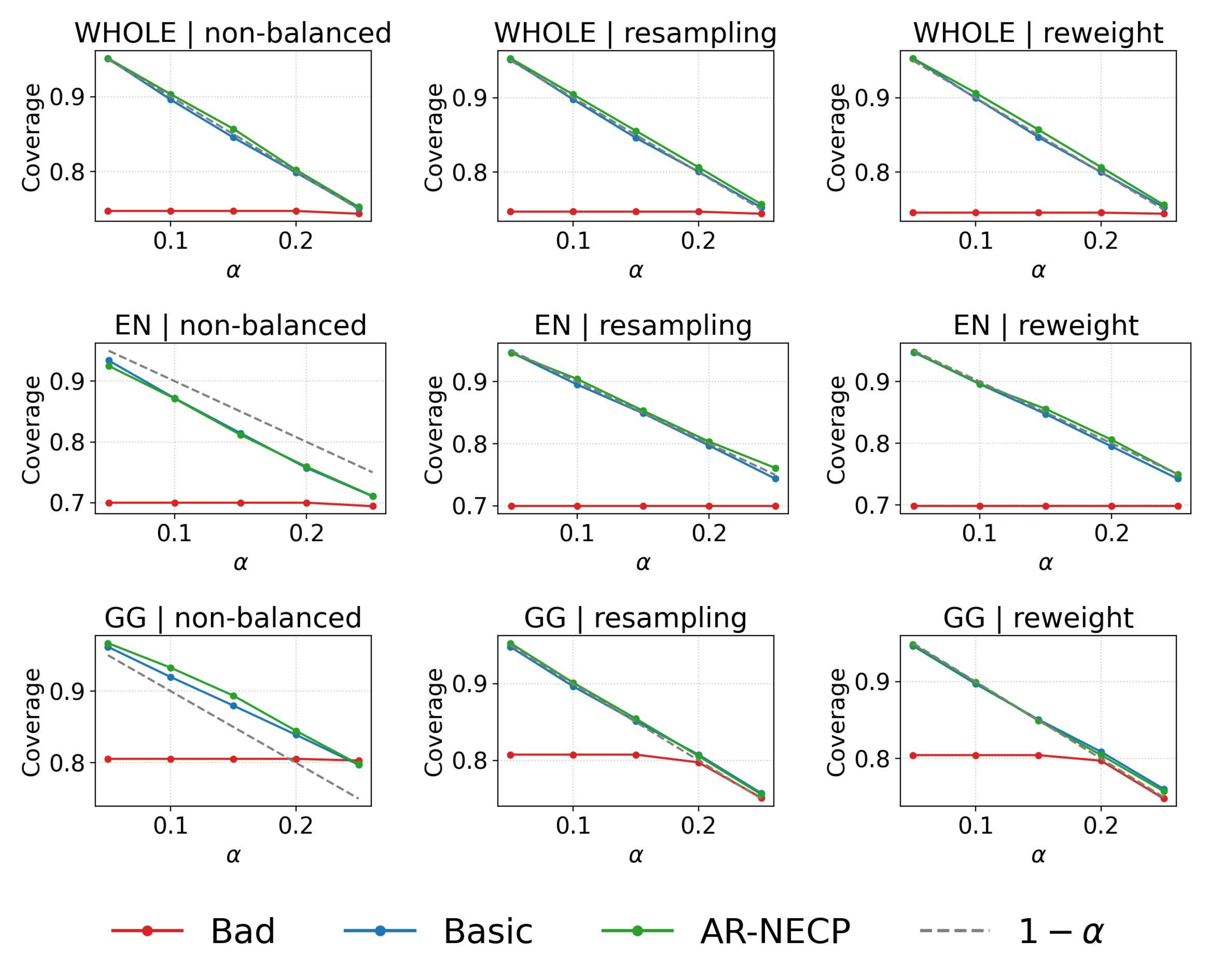}
    \caption{Coverage of the three CP methods under different quantiles, different distribution shifts, and using different balancing methods (resampling, reweighting, and non-balancing). Results that lie closest to the \(1-\alpha\) line indicate that the corresponding methods achieve the desired performance, neither exceeding it (overcoverage) nor falling short (undercoverage).}
    \label{fig:lc3}
\end{figure}
\subsection{Experiment Setting}
We evaluate \textbf{AR-NECP} on three widely used open QA datasets: TriviaQA \cite{2017arXivtriviaqa}, HotPotQA \cite{yang2018hotpotqa}, and MMLU \cite{hendryckstest2021, hendrycks2021ethics}. For each dataset, we first split it into multiple domains. We divide \textbf{TriviaQA} into five domains using keyword-based search. The corresponding keywords are: \textit{science \& art (sc)}, \textit{history (ht)}, \textit{sports \& politics (sp)}, \textit{entertainment (en)}, and \textit{geography (gg)}. For \textbf{HotPotQA}, we divide the data into three domains using keyword-based search: \textit{science \& art (sc)}, \textit{history (ht)}, \textit{entertainment \& geography (eg)}. For \textbf{MMLU}, we follow \citet{mmlu_auxiliary_trained_set} and divide the dataset into four domains: \textit{STEM (st)}, \textit{Humanities (hu)}, \textit{Social Science (ss)}, and \textit{Other (ot)}.

For each domain, we set the number of training samples $n_{\text{train}}^k = 10{,}000$, buffer samples $n_{\text{buffer}}^k = 2{,}000$, and both clustering and calibration samples $n_{\text{cluster}}^k = n_{\text{cal}}^k = 1{,}000$, with a unified calibration size $n_{\text{cal}}^{\text{union}} = 5{,}000$. The test size $n_{\text{test}}^k$ will be defined later when we discuss distribution shifts and how our method addresses them. Three LLMs are used: Mistral-7B-Instruct-v0.3 (Mistral7B) \cite{jiang2023mistral7b}, gemma-7b-it (gemma7b) \cite{team2024gemma}, and Llama-3.1-8B-Instruct (Llama8B) \cite{touvron2023llama}. Our discussion will be focused on results for Llama8B on TriviaQA, and results for the other two models can be found in the Appendix. We fine-tune each model using the hyperparameters provided in the Appendix \ref{train:appendix}.
We vary error levels $\alpha$ among [0.05, 0.10, 0.15, 0.20, 0.25].

We compare with two \textbf{baselines}: frequency-based CP (Bad) and basic “can’t answer” CP (Basic), adapting state-of-the-art methods from \cite{su-etal-2024-api} and \cite{li-etal-2024-traq} to the CDP setting. The former ignores LLMs' ability to abstain, while the latter introduces a “can’t answer” label but still fails to reject responses when the LLM cannot provide a valid answer. We apply the same data balancing techniques (for addressing non-exchangeability across domains) to all compared methods. Results are averaged from five runs.

\textbf{Metrics.}
We adopt three metrics to assess performance: (1) Coverage: the ratio of prediction sets that contain the ground truth. (2) Efficiency: the average size of the prediction sets. (3) Unanswerable efficiency: efficiency measured specifically on unanswerable questions, used to verify the effectiveness of our proposed methods. The first two metrics are standard in CP. 

\begin{figure}
    \centering
    \includegraphics[width=1\linewidth]{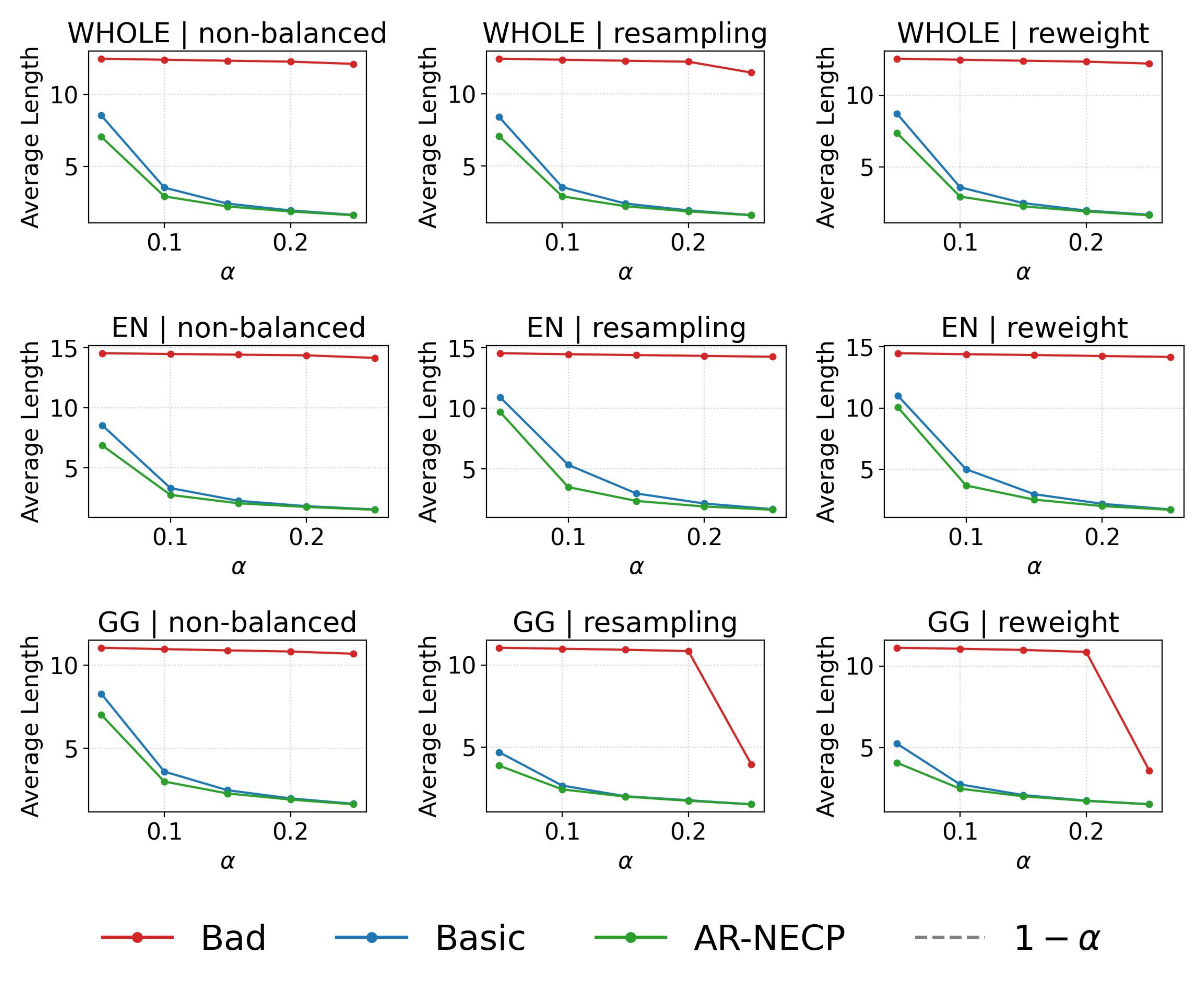}
    \caption{Average prediction set size of the three CP methods under different quantiles, under different distribution shifts, and using different balancing methods. A smaller value means better performance.}
    \label{fig:ll3}
\end{figure}
\subsection{Main Results (RQ1 \& RQ2)}
We evaluate coverage and efficiency across different values of $\alpha$ under three testing distribution settings:
(1) When easy questions dominate the test data, calibration causes over-coverage;
(2) When hard questions dominate, CP yields under-coverage; and
(3) When calibration and test distributions are similar, the target coverage is achieved.
For TriviaQA, we set 3000 questions in domain “gg” for the first case and 3000 in “en” for the second, with 200–500 questions randomly sampled for the remaining domains. In the balanced case, each domain contains 800–1500 randomly sampled questions. Results are shown in Figure~\ref{fig:lc3} and Figure~\ref{fig:ll3}.
We have the following observations:
\squishlist
\item The basic non-balanced CP method is sensitive to distribution shift. When the easier domain (GG) dominates the testing data, it tends to produce over-coverage; whereas when the harder domain (EN) dominates, it fails to reach the target coverage (the dotted line in Figure~\ref{fig:lc3}). This confirms that domain distribution indeed affects the coverage and highlights the necessity of addressing such distribution shifts.

\item Both resampling and reweighting methods produce prediction sets whose coverage closely matches the target level, empirically supporting that our approach successfully mitigates non-exchangeability under distribution shift. This also indicates that our estimation of domain proportions performs well in practice.

\item Our adaptive rejection CP consistently yields tighter prediction sets through its rejection mechanism, especially under high coverage requirements, demonstrating the robustness and effectiveness of the grid-search calibration. When the coverage level is low, the improvement becomes limited because the high inclusion threshold already filters out uncertain or unanswerable cases. 

\item The Bad CP baseline consistently shows lower efficiency than both the Basic CP and AR-NECP, and fails to achieve the target coverage in most scenarios, revealing the limitations of previous CP settings.
\squishend
\subsection{How do different UQ methods for LLM answer rejection influence prediction efficiency? (RQ3)}
\begin{table}[h]
\centering
\caption{Comparison of prediction efficiency for AR-NECP (MLP) and AR-NECP (NE) under reweighting across different domains distribution and $\alpha$ values.}
\resizebox{\linewidth}{!}{
\begin{tabular}{c|c|ccccc}
\hline
\textbf{Distribution} & \textbf{Method} & $\alpha$=0.05 & $\alpha$=0.10 & $\alpha$=0.15 & $\alpha$=0.20 & $\alpha$=0.25 \\
\hline
\multirow{2}{*}{whole} 
& AR-NECP (MLP)  & 8.66 & 3.60 & 2.47 & 1.92 & 1.65 \\
& AR-NECP (NE)   & 7.95 & 3.38 & 2.37 & 1.88 & 1.62 \\
\hline
\multirow{2}{*}{en-d} 
& AR-NECP (MLP)  & 11.07 & 4.51 & 3.06 & 2.24 & 1.79 \\
& AR-NECP (NE)   & 10.49 & 3.72 & 2.66 & 2.10 & 1.72 \\
\hline
\multirow{2}{*}{gg-d} 
& AR-NECP (MLP)  & 4.58 & 2.59 & 2.06 & 1.76 & 1.55 \\
& AR-NECP (NE)   & 4.29 & 2.54 & 2.04 & 1.75 & 1.54 \\
\hline
\end{tabular}}
\label{tab:re}
\end{table}

We compare two UQ approaches for LLM answer rejection: our proposed method, which uses Normalized Entropy (NE) to measure uncertainty, and a commonly used alternative that trains a small MLP classifier \cite{li-etal-2024-traq} on $D_{\text{cluster}}^{\text{union}}$---the union of all cluster sets $D_{\text{cluster}}^k, k \in [1, 2, \dots, K]$—to predict the LLM’s confidence in answering each question. For illustrative purposes, we evaluated both methods at five different quantile levels using the reweighting strategy. The comparison results, presented in Table~\ref{tab:re}, show that the NE-based AR-NECP consistently outperforms the MLP-based variant, demonstrating the superior ability of NE to distinguish between answerable and unanswerable questions. The smaller performance gap between AR-NECP (MLP) and AR-NECP (NE) in \textit{gg-d} domain is mainly because \textit{gg} is the most recent domain used to finetune LLMs in CDP, making its questions relatively easier for the LLM to answer; hence, the rejection mechanism has a marginal impact.
\subsection{How well does $\hat{n}^k_{\text{test}}$ work in practice? (RQ4)}
We examine whether estimating the number of test questions per domain is useful in real-world applications. 
For each testing distribution, we compare the estimated count $\hat{n}^k_{\text{test}}$ with the true count $n^k_{\text{test}}$ in every domain $k$ and report the mean relative error defined as: $\Delta^k \;=\; \frac{\left|\,n^k_{\text{test}}-\hat{n}^k_{\text{test}}\,\right|}{n^k_{\text{test}}}.$
Results are presented in Table~\ref{tab:mmd}. 
Overall, the errors are small across domains, which shows the estimator is accurate and useful in practice. 
The small bias comes from sampling variability; with more samples, for instance, en in en-d, gg in gg-d, the domain-wise I.I.D.\ assumption holds more closely and the estimates improve.

\begin{table}[h]
\centering
\resizebox{0.7\linewidth}{!}{
\begin{tabular}{c|ccccc}
\hline
\textbf{Shift} & \textbf{sc} & \textbf{ht} & \textbf{sp} & \textbf{en} & \textbf{gg} \\
\hline
en-d   & 0.08 & 0.04 & 0.05 & 0.01 & 0.10 \\
gg-d   & 0.10 & 0.11 & 0.11 & 0.06 & 0.02 \\
whole  & 0.04 & 0.02 & 0.03 & 0.02 & 0.04 \\
\hline
\end{tabular}
}
\caption{Mean relative error $\Delta^k$ between true and estimated domain counts under different test-time shifts. 
\textit{en-d} and \textit{gg-d} denote test sets dominated by the \textit{en} and \textit{gg} domains, respectively. Lower values is better.}
\label{tab:mmd}
\end{table}



\section{Conclusion}
This work takes an important step toward ensuring the reliability of self-evolving LLMs. We are the first to CDP where LLMs must continuously adapt to shifting knowledge domains. Our proposed AR-NECP framework addresses two fundamental barriers: handling unknown domain shifts between calibration and testing data, and enabling LLMs to selectively abstain when their competence varies across domains. By reweighting calibration data according to estimated domain distributions and incorporating a label-conditional rejection mechanism, AR-NECP delivers stronger statistical guarantees and more informative predictions. Extensive experiments verify its robustness and efficiency. Looking forward, this approach opens a pathway toward trustworthy CL, laying the foundation for LLMs that can evolve while maintaining reliability in real-world, dynamically changing environments.
\section*{Limitation}
There are several limitations in this work that warrant future research.
(1) Although our estimated number of test questions per domain is theoretically justified, in practice, it can still exhibit noticeable variance. How to accurately estimate and control this variance is beyond the scope of this paper and can be explored in future work.
(2) Another limitation lies in the need to resample calibration answers after each fine-tuning stage on a new dataset, which incurs additional computational cost. Ideally, calibration answers could be sampled once and reused in later stages to improve efficiency. However, this is not feasible because fine-tuning across domains introduces non-exchangeability among models trained at different steps. A potential direction for future research is to estimate and correct this bias by modeling how the number of training steps or the order of domain adaptation affects the calibration consistency.
\bibliography{custom}
\appendix

\section{Appendix}

\label{sec:appendix}
\subsection{Theoretical Proof}
\label{theory}

In this section, we provide theoretical proofs for the following statements:
\begin{enumerate}
    \item Our Adaptive Rejection CP can achieve better efficiency than standard CP with the `Can't answer' label, when the coverage requirement is high and the uncertainty score for identifying unanswerable questions is reasonable.
    \item Our Adaptive Rejection CP can achieve the required coverage.
\end{enumerate}
For our Non-exchangeable CP, we additionally prove:
\begin{enumerate}
    \item Weighted CP achieves global coverage.
    \item Resampling CP achieves asymptotic (infinite-sample) coverage.
\end{enumerate}

\paragraph{Efficiency of Adaptive Rejection CP.}
Our Adaptive Rejection CP (\textsc{AR-CP}) achieves higher efficiency than standard CP, particularly when the target coverage \(1-\alpha\) is large. 
Since \textsc{AR-CP} employs a grid search to find the pair \((\alpha_0, \alpha_1)\) that minimizes the expected prediction set size, we have the least-efficient baseline:
\begin{align}
E_{\alpha_0, \alpha_1}\!\big(|C^{\text{ARCP}}_\alpha(x_i)| \mid x_i \in D_{\text{test}}\big) \\\nonumber
= E_{\alpha}\!\big(|C_\alpha(x_i)| \mid x_i \in D_{\text{test}}\big),
\end{align}
when \(\alpha_1 = 0\) and \(\alpha_0 = \alpha\).
Thus, it suffices to show that for small \(\alpha\), there exist \((\alpha_0, \alpha_1)\) satisfying Eq.~\ref{a01} such that
\begin{align}
\label{t1}
E_{\alpha_0, \alpha_1}\!\big(|C^{\text{ARCP}}_\alpha(x_i)| \mid x_i \in D_{\text{test}}\big)\\\nonumber
<
E\!\big(|C_\alpha(x_i)| \mid x_i \in D_{\text{test}}\big).
\end{align}

We can express the expected prediction set size under \textsc{AR-CP} and $E\!\big(|C_\alpha(x_i)| \mid x_i \in D_{\text{test}}\big)$ as:
\begin{align}
&E_{\alpha_0, \alpha_1}\!\big(|C^{\text{ARCP}}_\alpha(x_i)| \mid x_i \in D_{\text{test}}\big) \nonumber\\
&= P\!\big(p^0(x_i)<q^0_{\alpha_0},p^1(x_i)<q^1_{\alpha_1}\big) \nonumber\\
&\quad + \big(E(|C^{ans}_\alpha(x_i)|)+\epsilon\big)
   P\!\big(p^1(x_i)<q^1_{\alpha_1}\big)\\ \nonumber
& <E(|C^{ans}_\alpha(x_i)|) + P\!\big(p^0(x_i)<q^0_{\alpha}\big),
\end{align}
where \(\epsilon \in \mathbb{R}\) is a small correction term used to mitigate calibration shift in \(\hat{q}_{\alpha}^{\text{text}}\). $E(|C^{ans}_\alpha(x_i)|)$ is the average prediction set size of answers without the `Can't answer' label.

set $t=P\!\big(p^0(x_i)<q^0_{\alpha_0}\big)-P\!\big(p^0(x_i)<q^0_{\alpha}\big)>0$ which will be close to $\alpha-\alpha_0$ when uncertainty measurement performs well.
we can get Eq. \ref{t1} stand when:
\begin{align}
\label{t12}
&t+(E(|C_\alpha(x_i)|\mid x_i \in D_{\text{test}})+\epsilon)P(p^1(x_i)<q^1_{\alpha_1})\\\nonumber
&< E(|C_\alpha(x_i)|\mid x_i \in D_{\text{test}})
\end{align}
Let $A := \mathbb{E}(|C_\alpha(x_i)| \mid x_i\in D_{\text{test}})$,
$p := \mathbb{P}(p^{1}(x_i) < q^{1}_{\alpha_1})$,
and $t := \mathbb{P}(p^{0}(x_i) < q^{0}_{\alpha_0}) - \mathbb{P}(p^{0}(x_i) < q^{0}_{\alpha})$.
Then inequality Eq. \ref{t1} holds if and only if
\begin{align}
\frac{t + \epsilon\, p}{1-p} < A
\quad\Longleftrightarrow\quad
p < \frac{A - t}{A + \epsilon},
\end{align}
provided that $0 \le p < 1$ and $t < A$.
Since $p$ is monotone in $\alpha_1$, there exists
\[
\alpha_1^\star := \inf\big\{\alpha_1:\; \mathbb{P}(p^{1}(x_i) < q^{1}_{\alpha_1}) \le (A - t)/(A + \epsilon)\big\},
\]
such that Eq. \ref{t1} holds for all $\alpha_1 \ge \alpha_1^\star$.

Here $t \approx \alpha - \alpha_0$, so a smaller $\alpha_0$ (i.e., stricter rejection threshold)
increases $t$ and tightens the bound $(A - t)/(A + \epsilon)$,
requiring a smaller $p$ to satisfy~(31).
Since $p = \mathbb{P}(p^1(x_i) < q^1_{\alpha_1})$ decreases with larger $\alpha_1$,
There exists a critical value $\alpha_1^\star$ beyond which
The inequality naturally holds.
In practice, $t < A$ almost always holds when $\alpha$ is small because $A$---the expected prediction set size---is typically much larger than $\alpha - \alpha_0$.


\paragraph{Coverage Guarantee of Weighted CP.}
Under the domain I.I.D.\ assumption and the explanation in Section~\ref{samplling}, our estimated \(\hat{n}_{\text{test}}^k\) provides a valid approximation of the domain proportions in the testing data~\cite{tibshirani2019conformal}. 
The coverage guarantee of Weighted CP under covariate shift follows directly from prior work~\cite{tibshirani2019conformal}. 
By reweighting calibration samples according to the density ratio between the test and calibration distributions,
\begin{align}
w(x) = \frac{p_{\text{test}}(x)}{p_{\text{cal}}(x)},
\end{align}
the resulting weighted conformal predictor satisfies the asymptotic coverage guarantee:
\begin{align}
\mathbb{P}_{(x,y)\sim P_{\text{test}}}\!\big(y \in C_\alpha(x)\big) 
\ge 1 - \alpha.
\end{align}

\paragraph{Coverage Guarantee of Resampled CP.}
Under the domain I.I.D.\ assumption, when the number of resampled calibration samples is sufficiently large, the resampled calibration data approximates the test domain distribution. 
Hence, calibration and test sets can be regarded as I.I.D., guaranteeing that the prediction set achieves the desired coverage.

\subsection{Keywords for dividing datasets into different domains}
We categorize the TriviaQA dataset into five domains according to distinctive lexical cues.
Science \& Art includes a broad range of scientific and artistic terminology, encompassing words such as physics, chemistry, biology, scientist, astronomy, mathematics, theory, planet, galaxy, telescope, microscope, quantum, molecule, enzyme, neuron, genome, DNA, equation, calculus, genetics, and biochemistry, along with artistic and literary terms like poet, poem, novel, novelist, painter, composer, symphony, art, gallery, museum, playwright, theatre, literature, ballet, opera, sonnet, sculpture, poetry, and sculptor.
History includes keywords referring to historical eras, figures, and events, including dynasty, emperor, treaty, revolution, war, ancient, medieval, empire, king, queen, monarch, colonial, renaissance, pharaoh, roman, greek, battle, republic, treaties, and monarchy.
Sports \& Politics merges athletic and political expressions, including olympics, fifa, uefa, nba, mlb, nfl, nhl, footballer, coach, tournament, world cup, premier league, grand slam, stadium, athlete, league, match, finals, medal, goal, club, referee, umpire, arena, president, prime minister, parliament, election, constitution, senate, cabinet, ministry, chancellor, congress, referendum, party, coalition, diplomacy, treaties, regime, governor, and minister.
Entertainment focuses on mass media and performing arts, including film, movie, actor, actress, TV, television, oscars, academy award, grammy, emmy, director, screenplay, soundtrack, series, episode, band, album, song, box office, festival, orchestra, opera, musical, and concert.
Finally, Geography covers geographic and geopolitical terminology such as capital, river, mount, mountain, province, city, country, island, lake, ocean, sea, peninsula, desert, continent, bay, harbor, border, valley, delta, latitude, longitude, archipelago, and coast.

We categorize the HotpotQA dataset into three semantic domains based on representative keywords.
Science \& Art contains terms that reflect both scientific and artistic concepts, including physics, chemistry, biology, scientist, astronomy, mathematics, theory, planet, galaxy, telescope, microscope, quantum, enzyme, neuron, genome, DNA, equation, calculus, genetics, biochemistry, poet, poem, painter, composer, symphony, gallery, playwright, ballet, opera, sculpture, poetry, and sculptor.

History focuses on terminology related to historical events, figures, and periods, including dynasty, emperor, treaty, revolution, war, ancient, medieval, empire, king, queen, monarch, colonial, renaissance, pharaoh, roman, greek, battle, republic, treaties, and monarchy.
Entertainment \& Geography combines cultural and locational entities, covering entertainment-related words such as film, movie, actor, actress, TV, television, oscars, academy award, grammy, emmy, director, screenplay, soundtrack, series, episode, band, album, song, box office, festival, orchestra, opera, musical, concert, and novel, as well as geographical expressions like capital, river, mount, mountain, province, city, country, island, lake, ocean, sea, peninsula, desert, continent, bay, harbor, border, valley, delta, latitude, and longitude.
\subsection{Training and sampling detail}
\label{train:appendix}
The learning rate is set to \(1\times10^{-4}\), and the model is trained for three epochs at each step. 
During answer sampling, we use a temperature of 1.2, top-\(p=0.9\), top-\(k=100\), and a repetition penalty of 1.0.

\subsection{Additional experiment on Main (RQ1 \& RQ2)}
\label{main:appendix}
\begin{figure}
    \centering
    \includegraphics[width=1\linewidth]{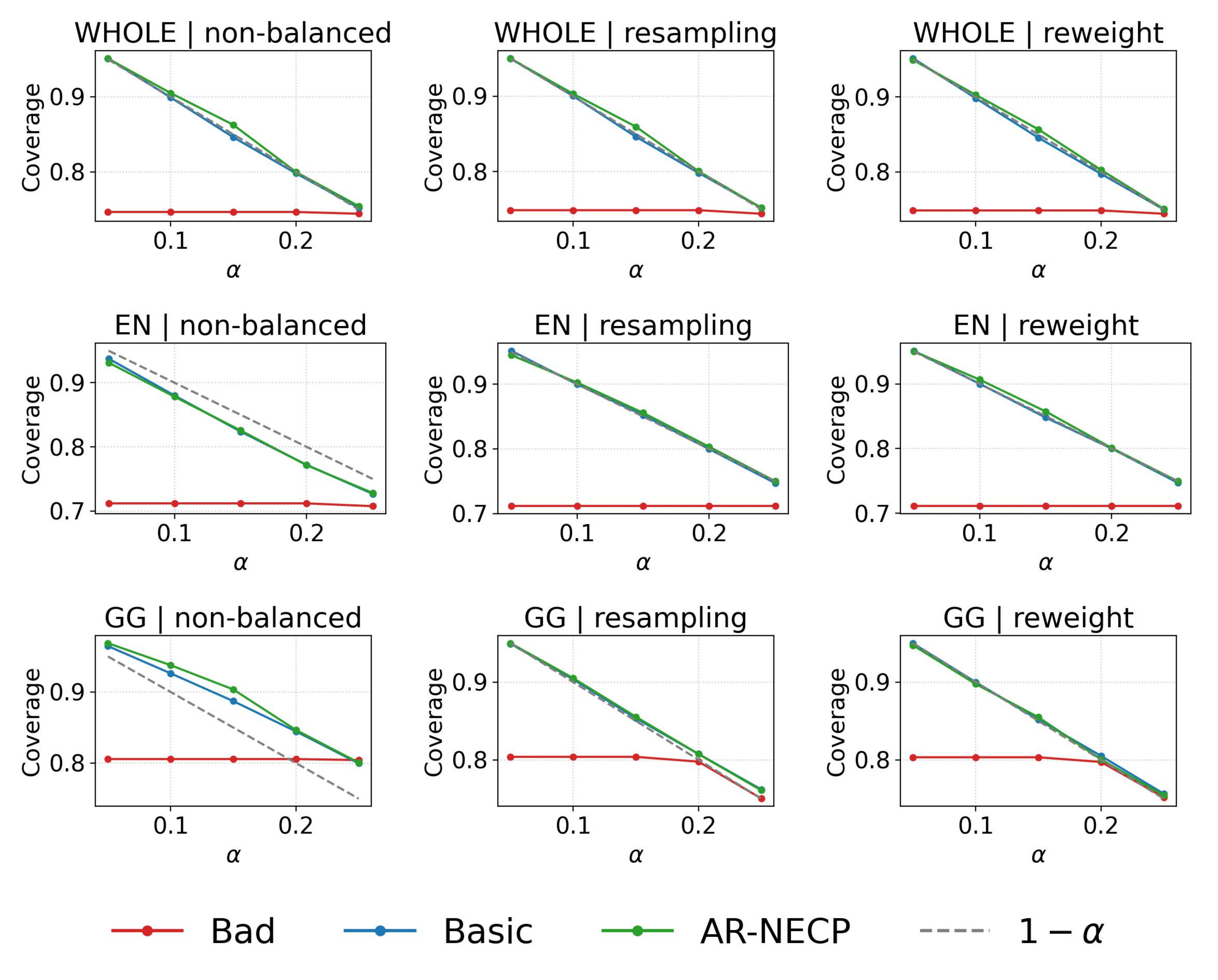}
    \caption{Coverage of the three CP methods under different quantiles under different distribution shifts and using different balancing methods. Results for Mistral7B on TriviaQA.}
    \label{fig:mc3}
\end{figure}

\begin{figure}
    \centering
    \includegraphics[width=1\linewidth]{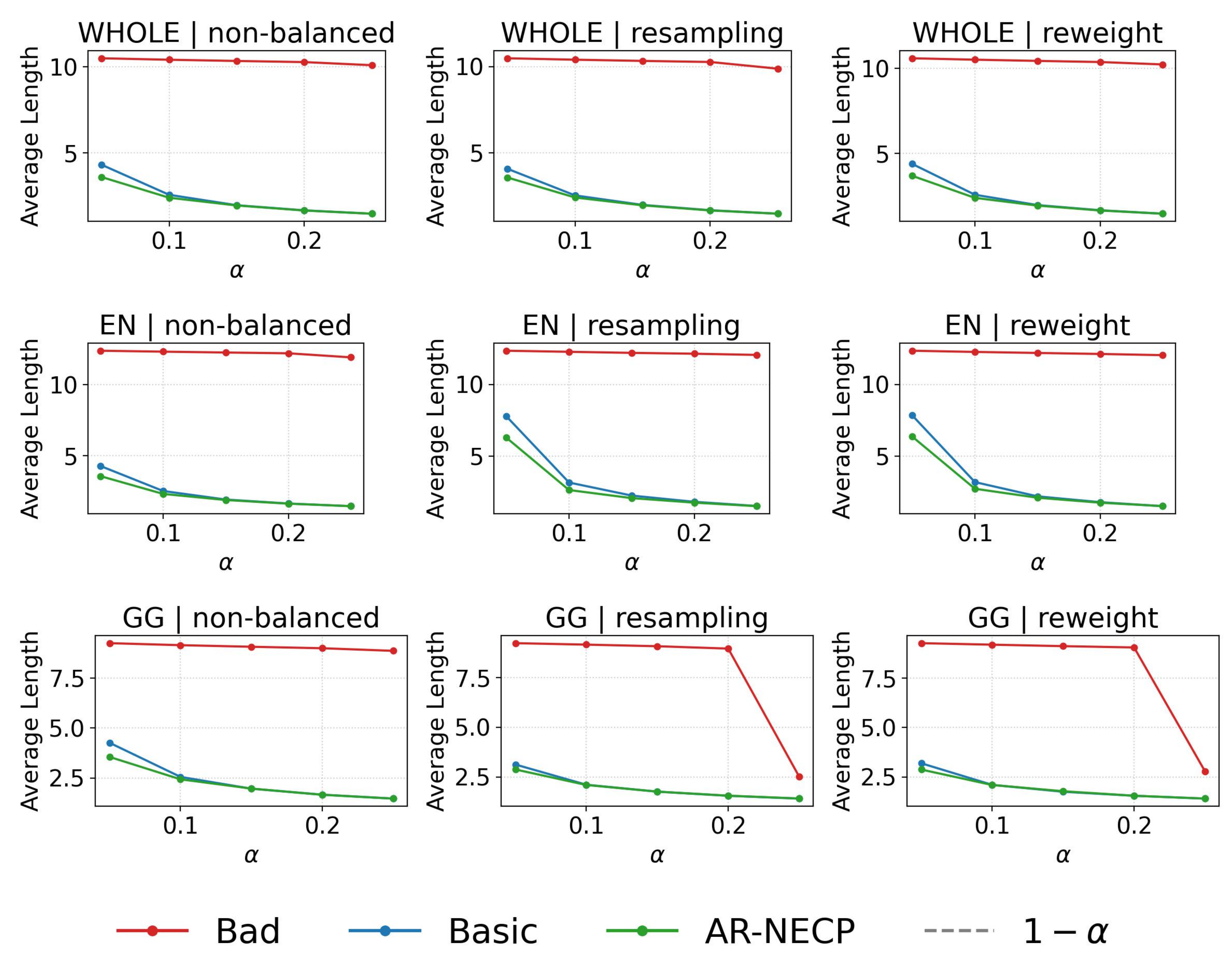}
    \caption{Average prediction set size of the three CP methods under different quantiles, under different distribution shifts, and using different balancing methods. Results for Mistral7B on TriviaQA.}
    \label{fig:ml3}
\end{figure}

\begin{figure}
    \centering
    \includegraphics[width=1\linewidth]{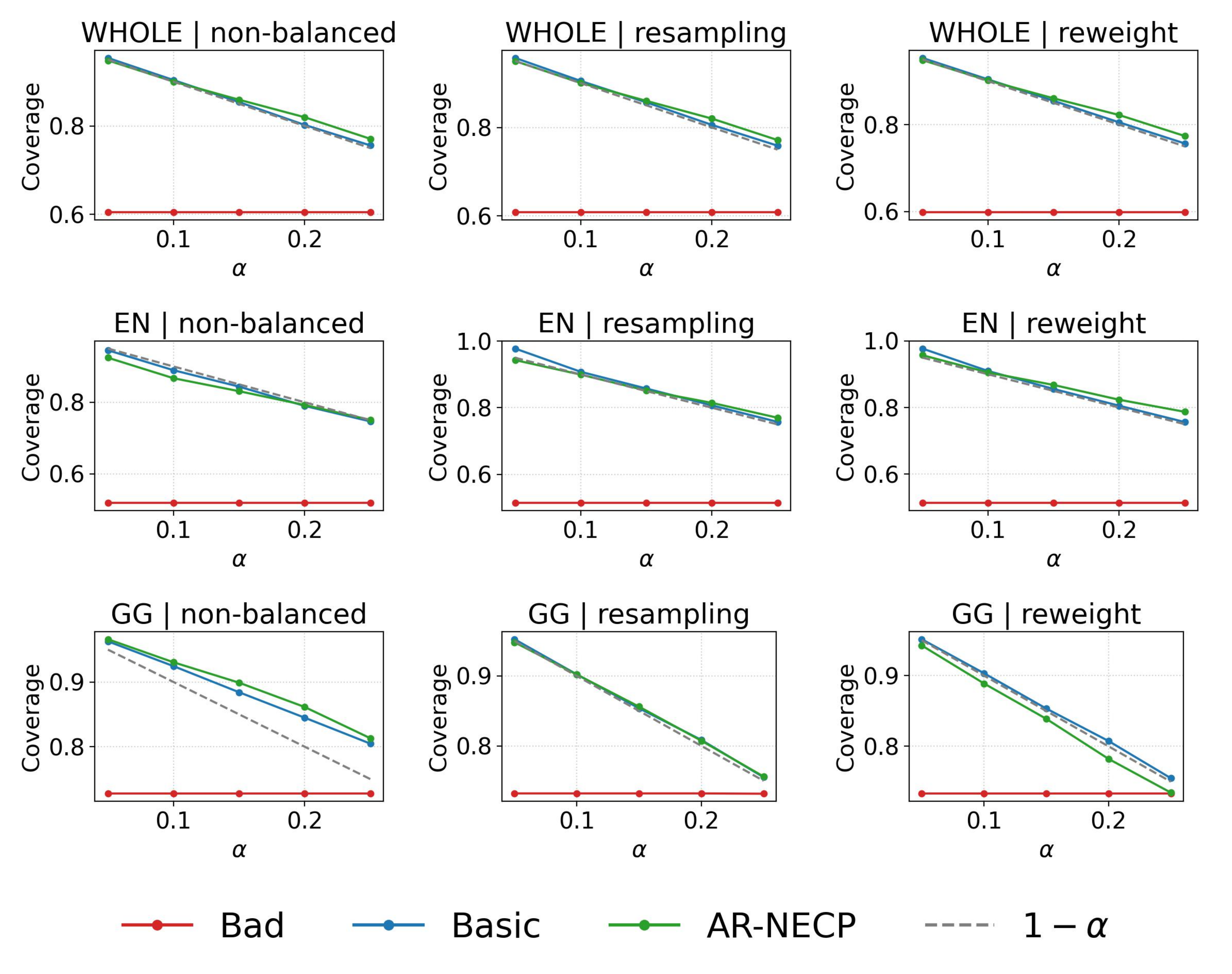}
    \caption{Coverage of the three CP methods under different quantiles under different distribution shifts and using different balancing methods. Results for Gemma7B on TriviaQA.}
    \label{fig:gc3}
\end{figure}

\begin{figure}
    \centering
    \includegraphics[width=1\linewidth]{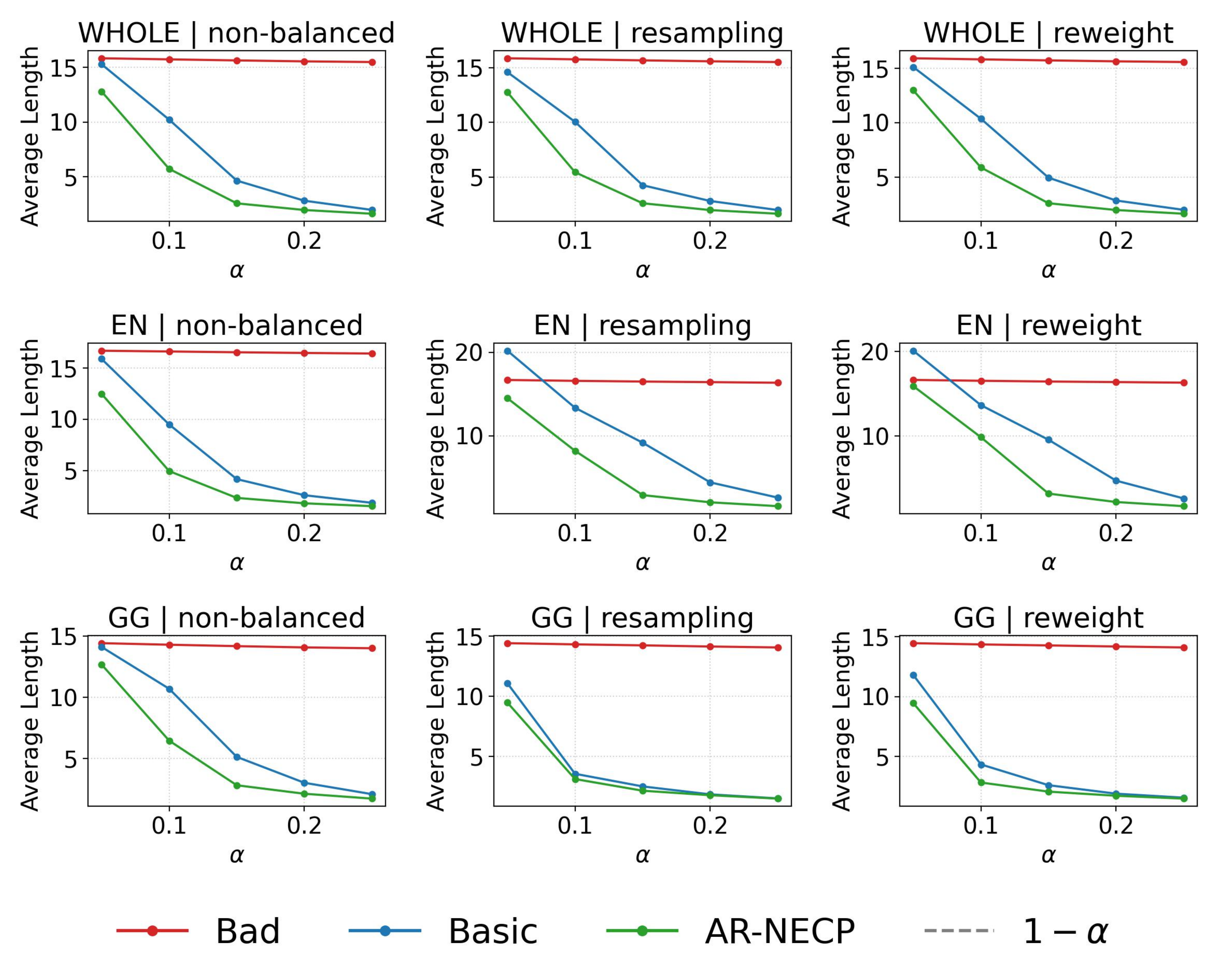}
    \caption{Average prediction set size of the three CP methods under different quantiles, under different distribution shifts, and using different balancing methods. Results for Gemma7B on TriviaQA.}
    \label{fig:gl3}
\end{figure}

\begin{figure}
    \centering
    \includegraphics[width=1\linewidth]{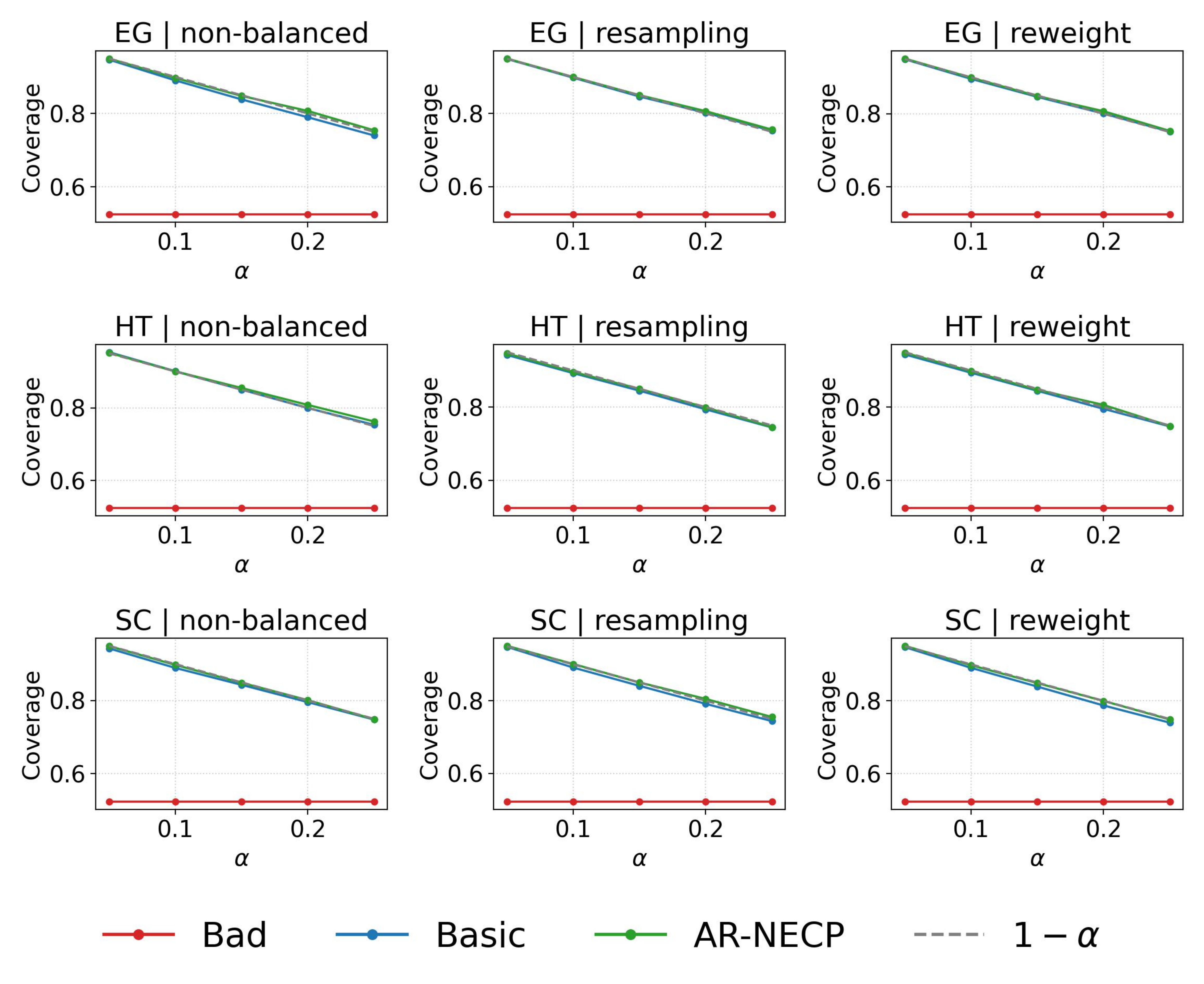}
    \caption{Coverage of the three CP methods under different quantiles under different distribution shifts and using different balancing methods. Results for Mistral7B on HotpotQA.}
    \label{fig:hmc3}
\end{figure}

\begin{figure}
    \centering
    \includegraphics[width=1\linewidth]{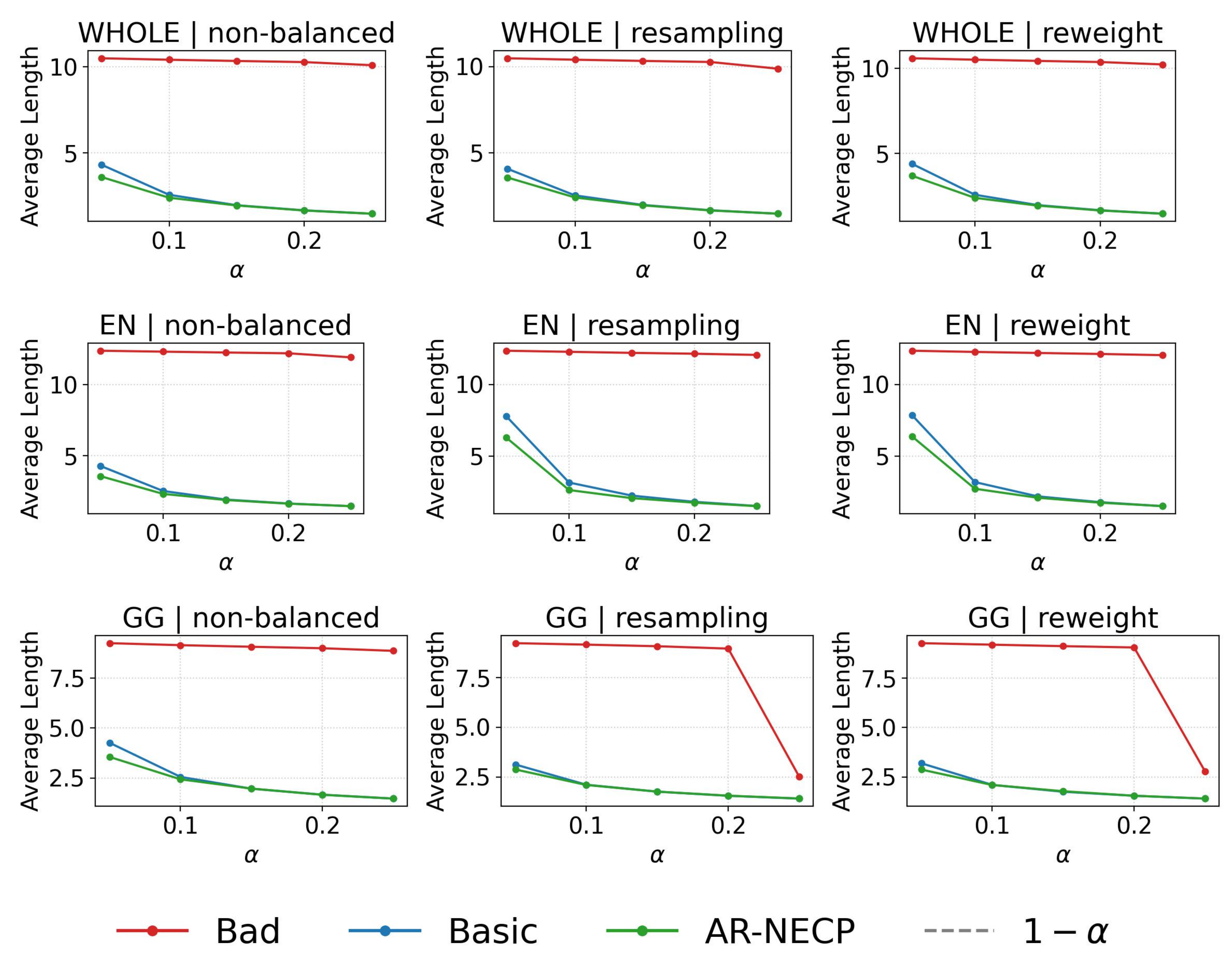}
    \caption{Average prediction set size of the three CP methods under different quantiles, under different distribution shifts, and using different balancing methods. Results for Mistral7B on HotpotQA.}
    \label{fig:hml3}
\end{figure}

\begin{figure}
    \centering
    \includegraphics[width=1\linewidth]{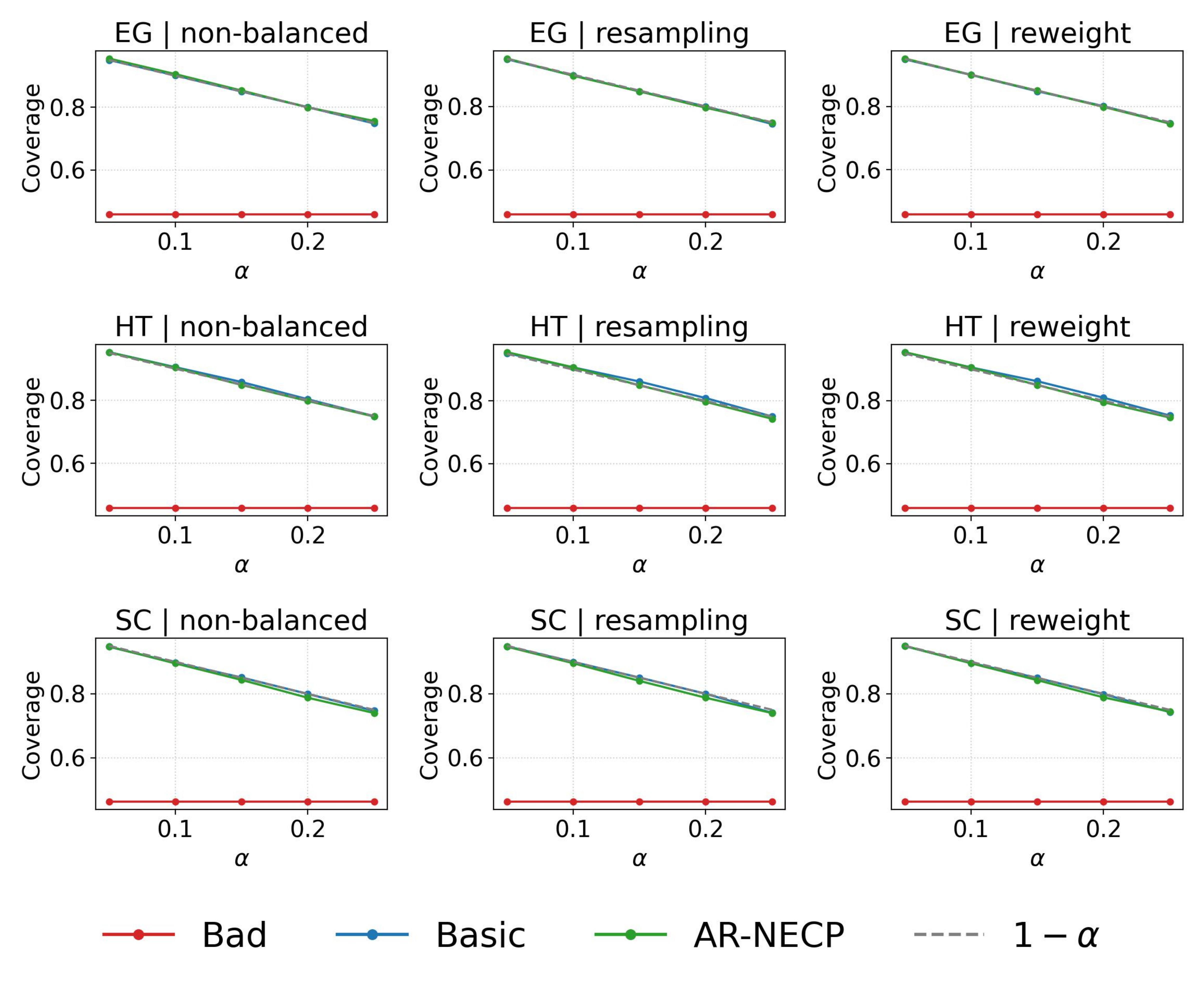}
    \caption{Coverage of the three CP methods under different quantiles under different distribution shifts and using different balancing methods. Results for Gemma7B on HotpotQA.}
    \label{fig:hgc3}
\end{figure}

\begin{figure}
    \centering
    \includegraphics[width=1\linewidth]{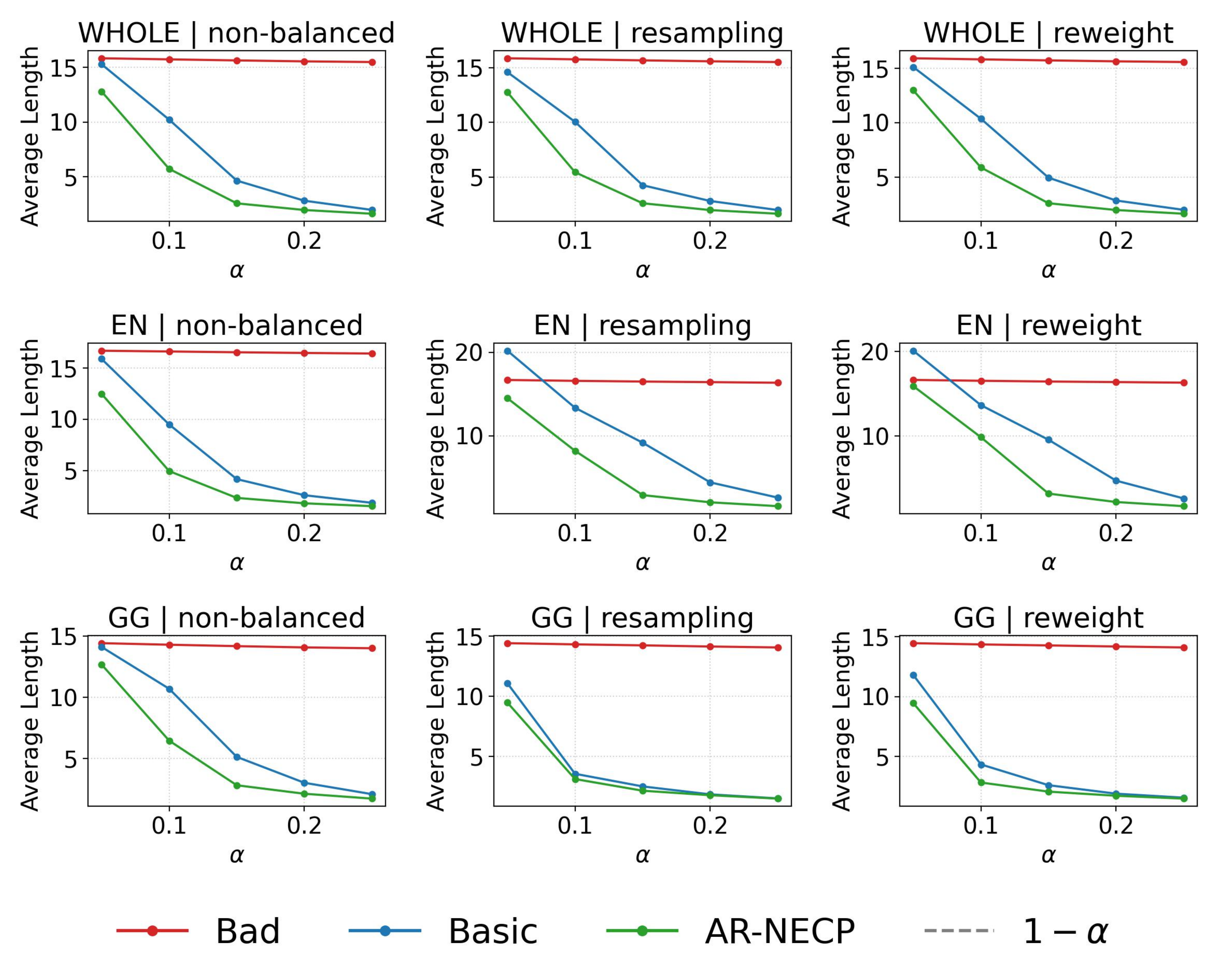}
    \caption{Average prediction set size of the three CP methods under different quantiles, under different distribution shifts, and using different balancing methods. Results for Gemma7B on HotpotQA.}
    \label{fig:hgl3}
\end{figure}

\begin{figure}
    \centering
    \includegraphics[width=1\linewidth]{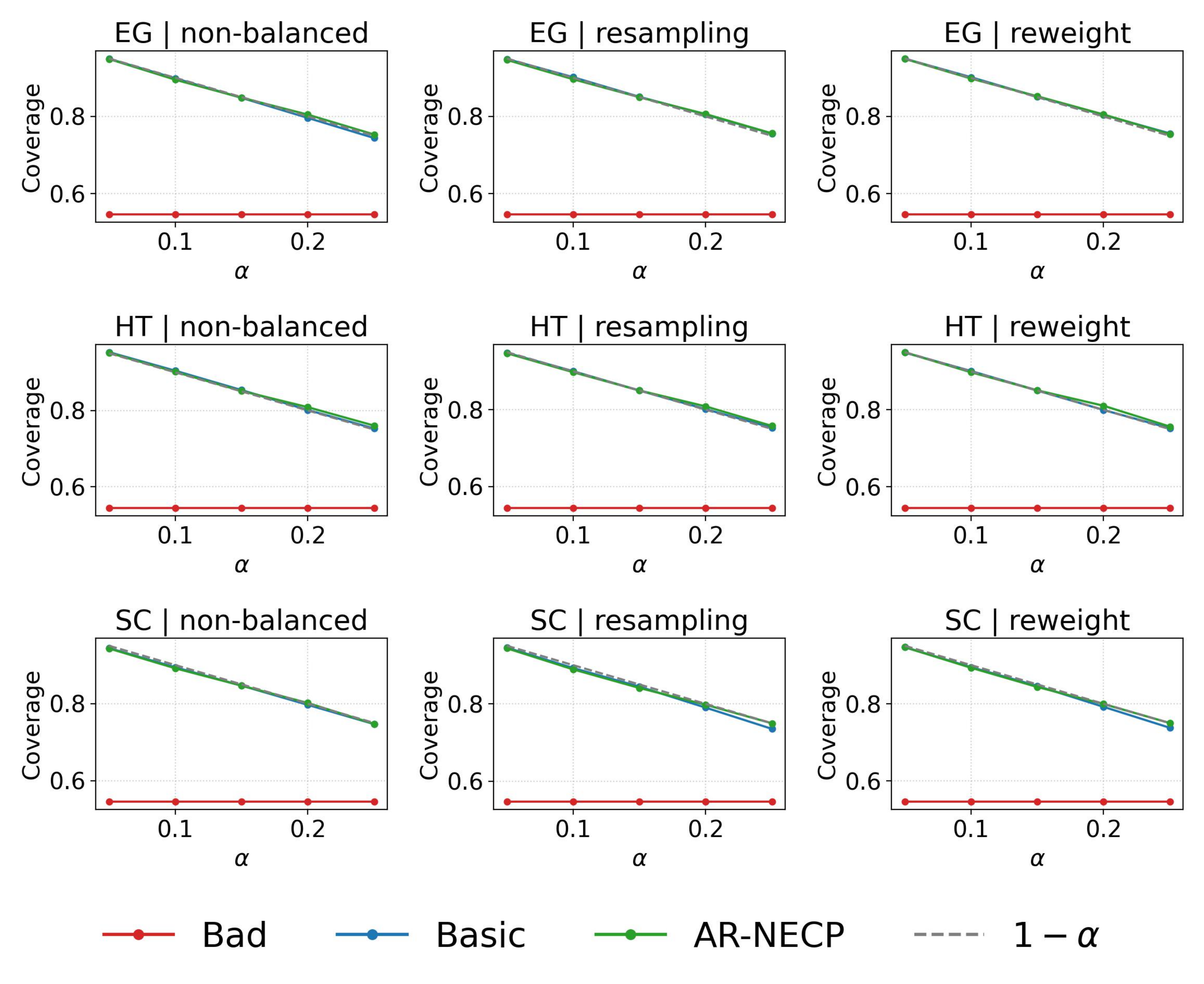}
    \caption{Coverage of the three CP methods under different quantiles under different distribution shifts and using different balancing methods. Results for Llama8B on HotpotQA.}
    \label{fig:hlc3}
\end{figure}

\begin{figure}
    \centering
    \includegraphics[width=1\linewidth]{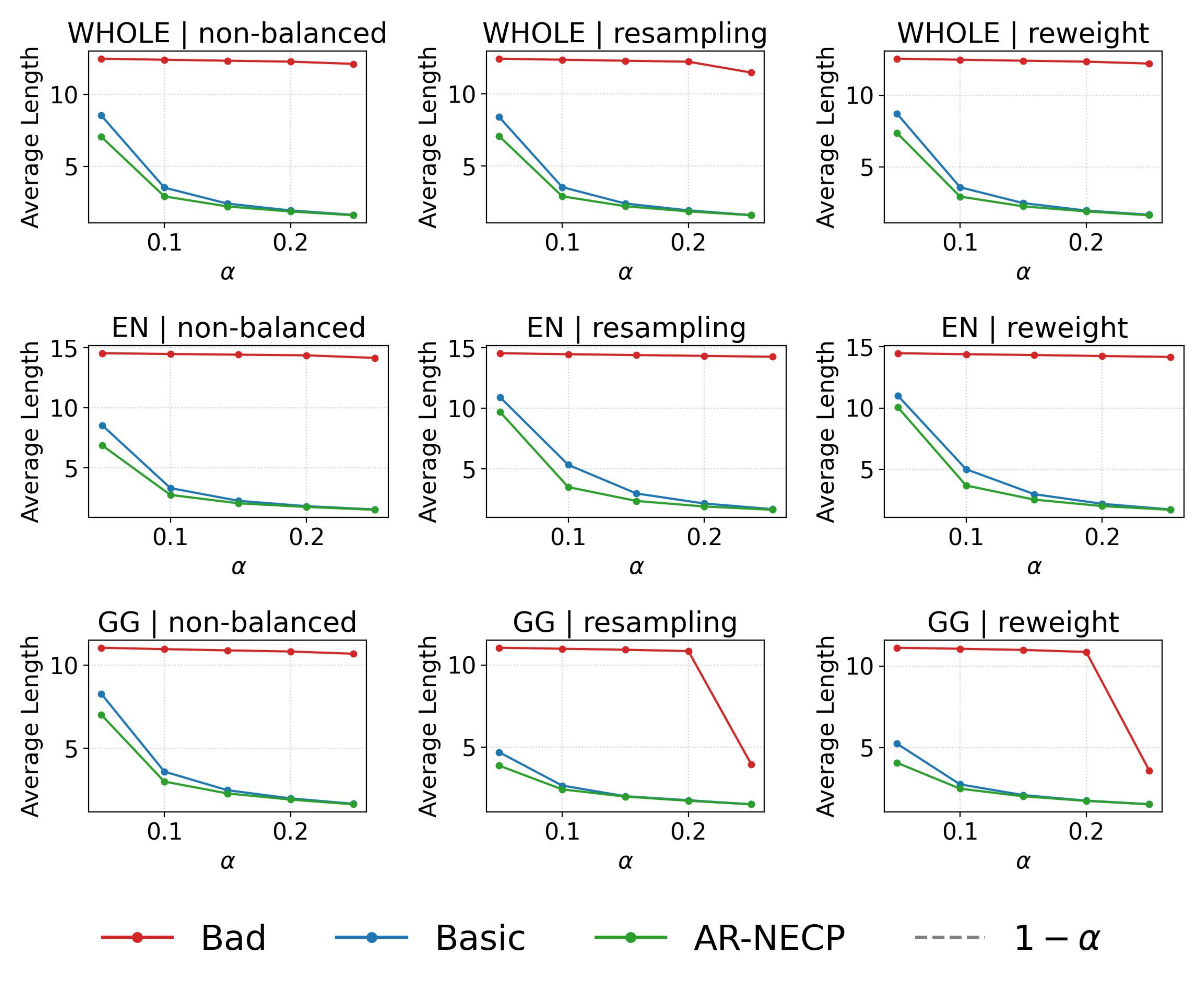}
    \caption{Average prediction set size of the three CP methods under different quantiles, under different distribution shifts, and using different balancing methods. Results for Llama8B on HotpotQA.}
    \label{fig:hll3}
\end{figure}

The results of Gemma7B and Mistral7B are shown in Figure \ref{fig:mc3}, \ref{fig:gl3}, \ref{fig:ml3}, \ref{fig:gc3}. They show the same finding as the main text. 
The results of Llama8B, Gemma7B, and Mistral7B on HotpotQA are shown in Figures~\ref{fig:hmc3}, \ref{fig:hgl3}, \ref{fig:hml3}, \ref{fig:hlc3}, \ref{fig:hll3}, and \ref{fig:hgc3}. 
It is worth noting that HotpotQA is a multi-hop reasoning dataset, which makes it challenging for LLMs to fine-tune effectively. 
This harder learning problem tends to make the model more resistant to catastrophic forgetting, resulting in minor shifts in the non-conformity score distribution. 
Therefore, in this dataset, we mainly focus on evaluating the performance of our Adaptive Rejection CP.


\begin{figure}
    \centering
    \includegraphics[width=1\linewidth]{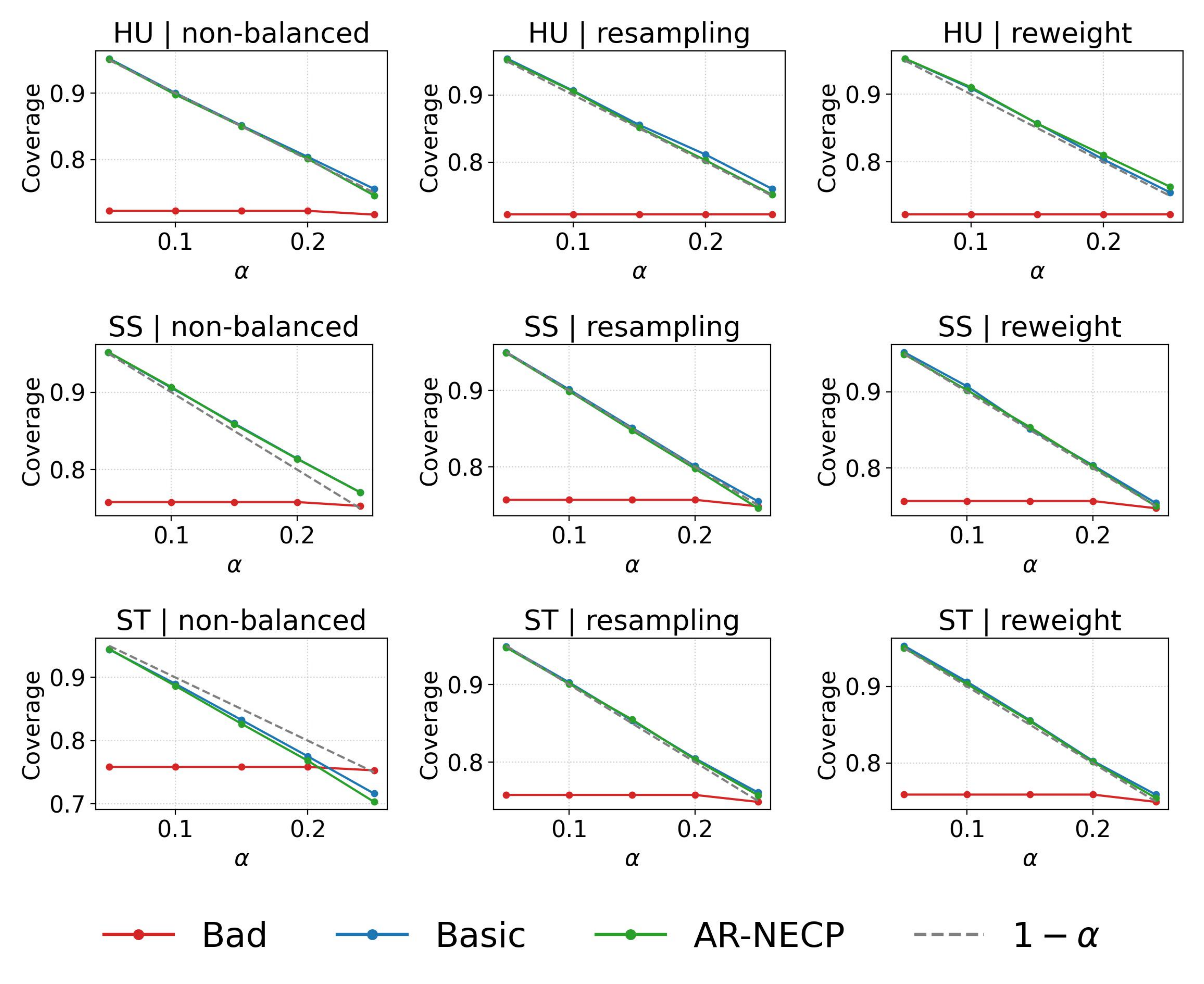}
    \caption{Coverage of the three CP methods under different quantiles under different distribution shifts and using different balancing methods. Results for Mistral7B on MMLU.}
    \label{fig:mmc3}
\end{figure}

\begin{figure}
    \centering
    \includegraphics[width=1\linewidth]{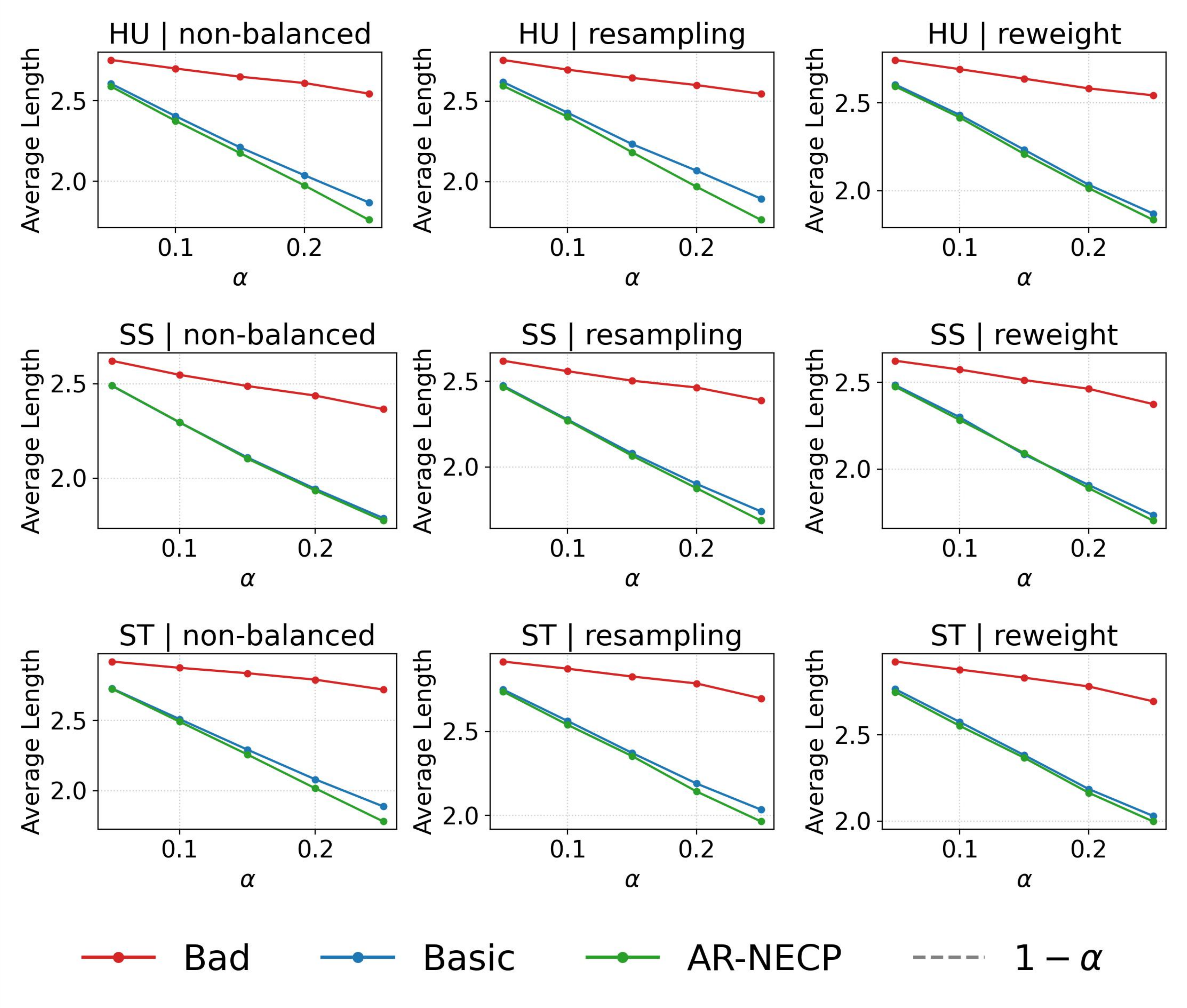}
    \caption{Average prediction set size of the three CP methods under different quantiles, under different distribution shifts, and using different balancing methods. Results for Mistral7B on MMLU.}
    \label{fig:mml3}
\end{figure}

\begin{figure}
    \centering
    \includegraphics[width=1\linewidth]{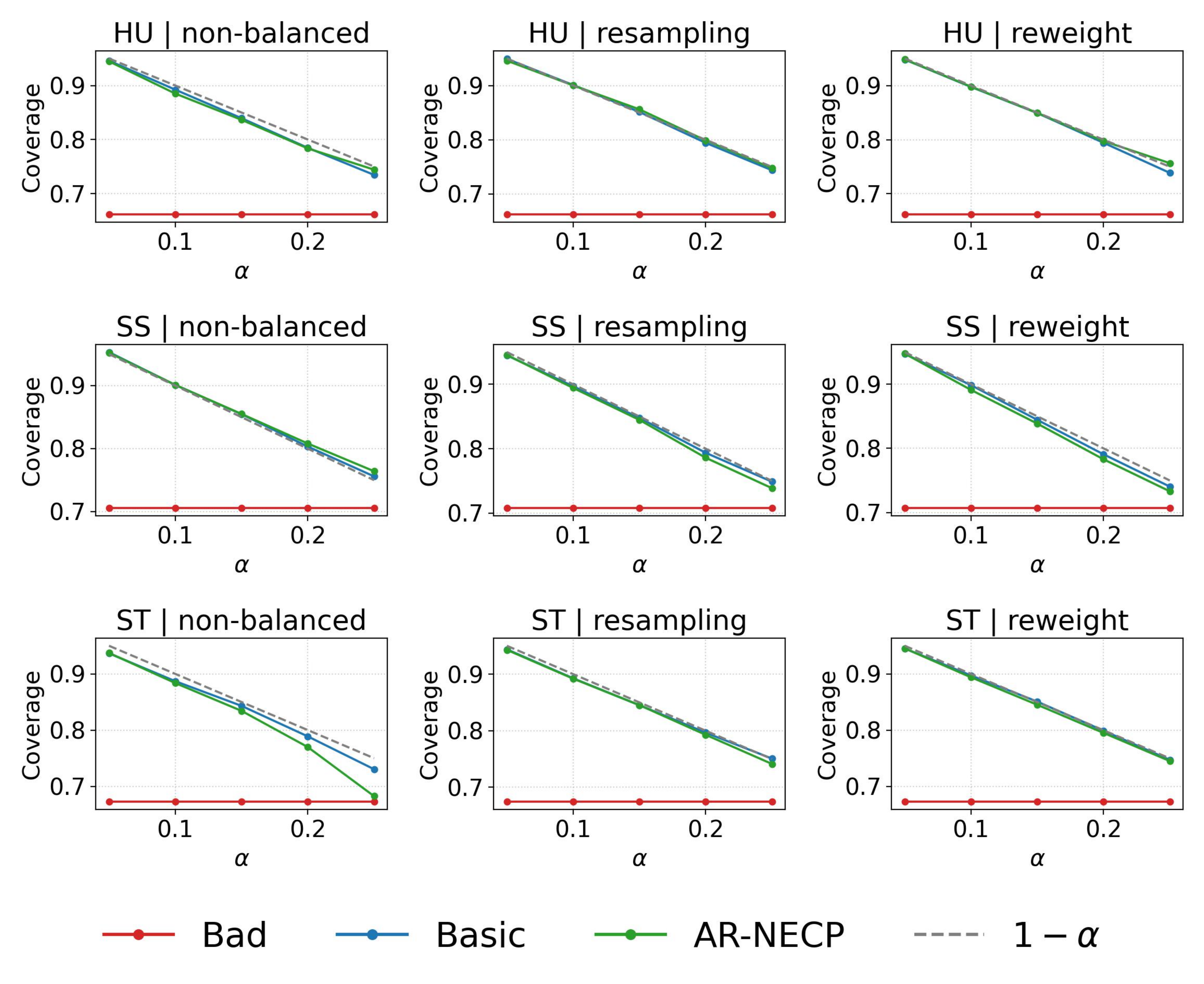}
    \caption{Coverage of the three CP methods under different quantiles under different distribution shifts and using different balancing methods. Results for Gemma7B on MMLU.}
    \label{fig:mgc3}
\end{figure}

\begin{figure}
    \centering
    \includegraphics[width=1\linewidth]{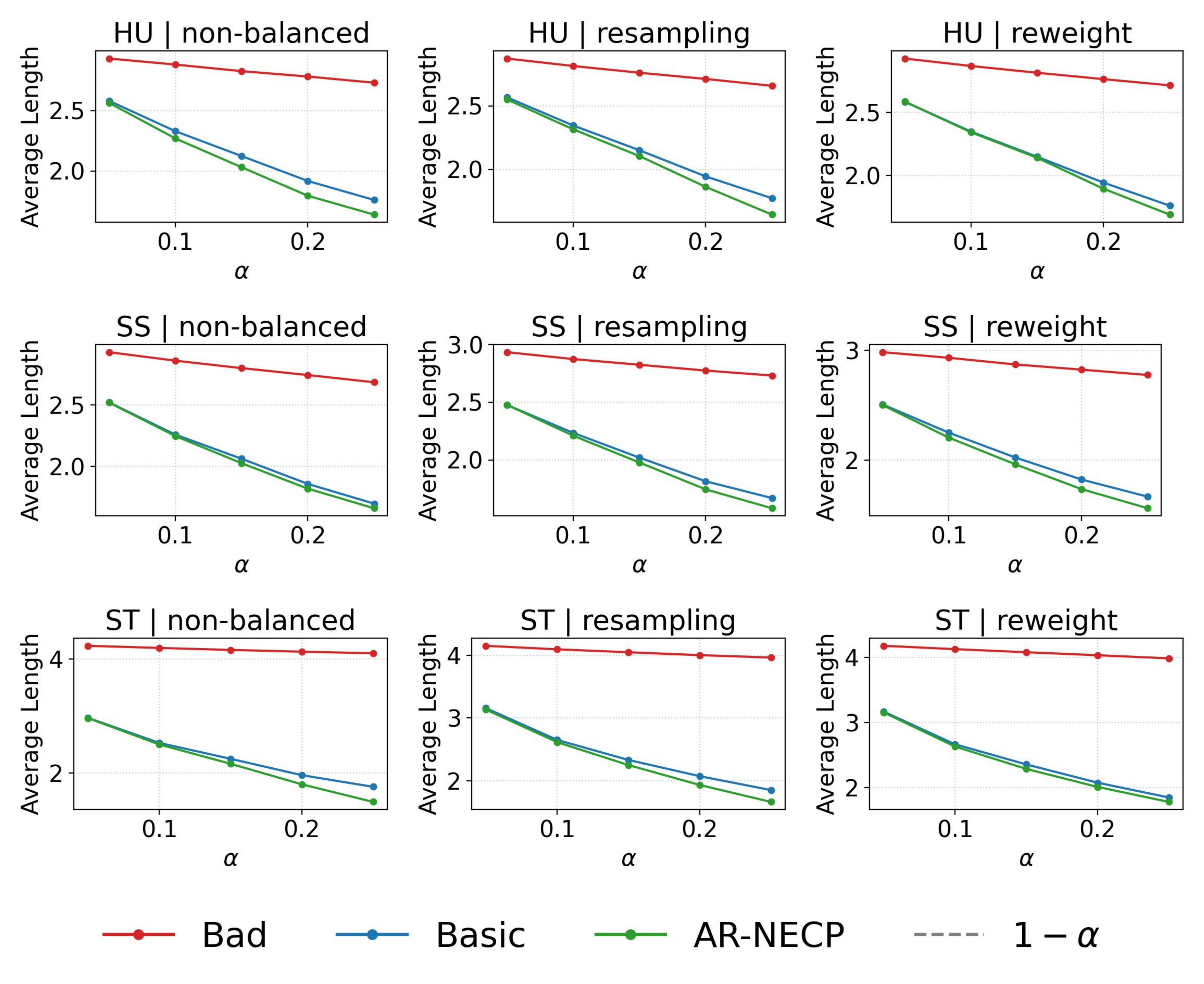}
    \caption{Average prediction set size of the three CP methods under different quantiles, under different distribution shifts, and using different balancing methods. Results for Gemma7B on MMLU.}
    \label{fig:mgl3}
\end{figure}

\begin{figure}
    \centering
    \includegraphics[width=1\linewidth]{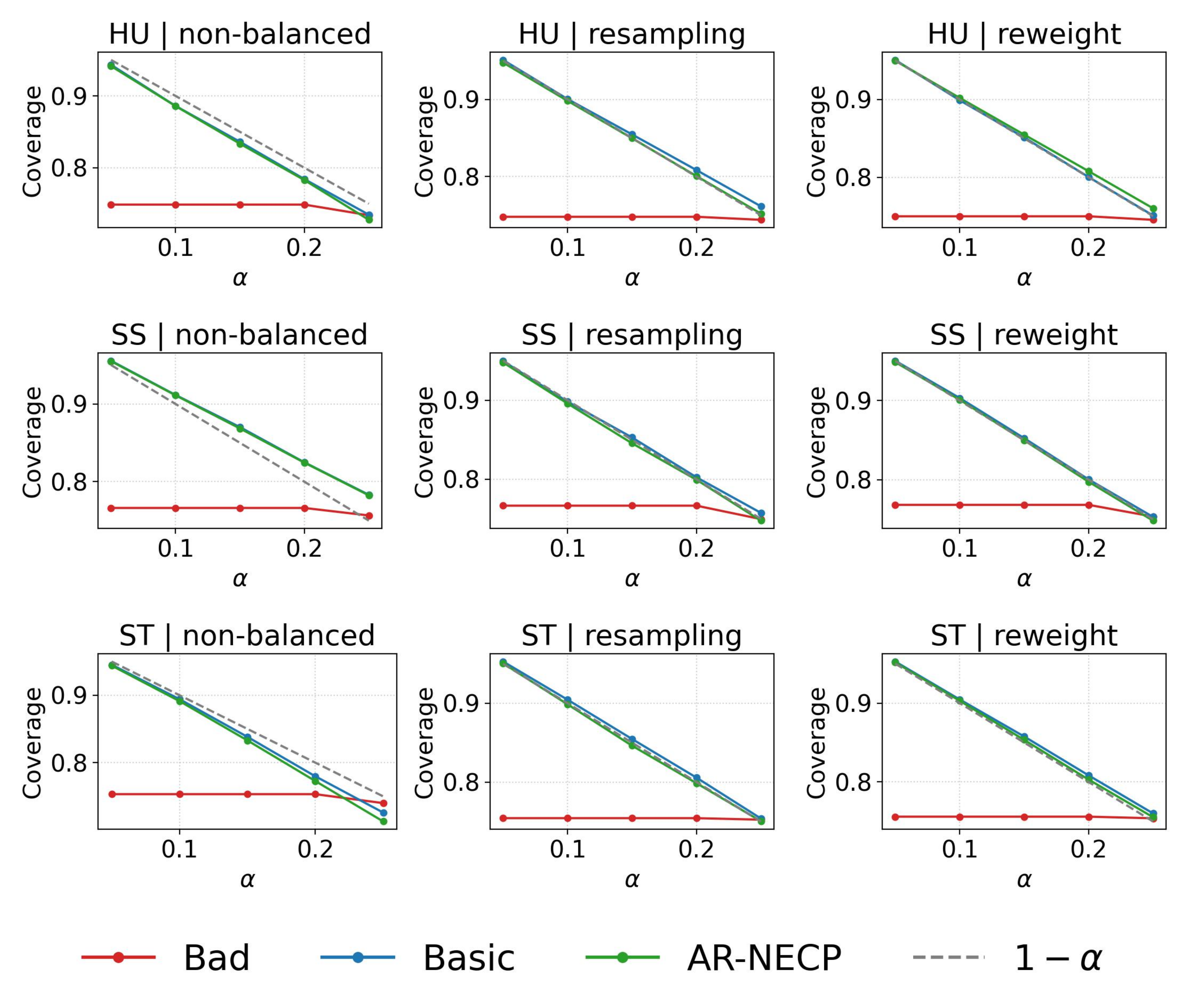}
    \caption{Coverage of the three CP methods under different quantiles under different distribution shifts and using different balancing methods. Results for Llama8B on MMLU.}
    \label{fig:mlc3}
\end{figure}

\begin{figure}
    \centering
    \includegraphics[width=1\linewidth]{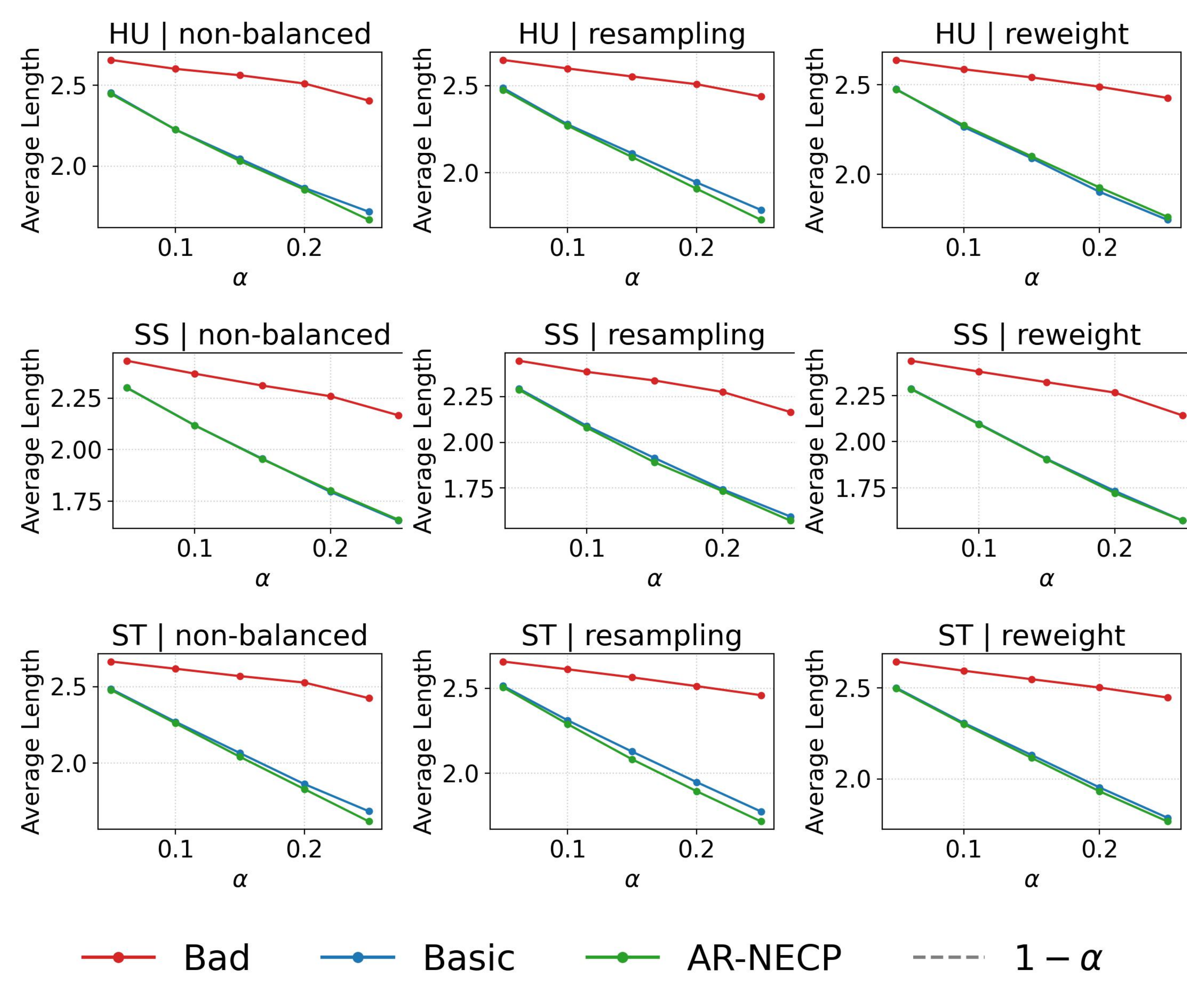}
    \caption{Average prediction set size of the three CP methods under different quantiles, under different distribution shifts, and using different balancing methods. Results for Llama8B on MMLU.}
    \label{fig:mll3}
\end{figure}

In the MMLU dataset, which is a multiple-choice QA benchmark, it is relatively easier for large language models to obtain correct answers through sampling. Therefore, we mainly focus on evaluating our method’s effectiveness in addressing distribution shifts. For efficiency analysis, the coverage results are presented in Figure \ref{fig:mmc3}, \ref{fig:mgc3}, \ref{fig:mlc3}, \ref{fig:mml3}, \ref{fig:mgl3}, \ref{fig:mll3}. From these results, we observe that our method effectively mitigates the distribution shift, while also improving efficiency. However, the improvement in efficiency is limited, as the task itself is relatively simple—the model only needs to choose from four possible answers.


\subsection{Additional experiment on RQ3}
Here are the results from TriviaQA on the model Mistral7B and gemma-7b-it. Under most settings, NE performs better than MLP.
\begin{table}[h]
\centering
\caption{Comparison of prediction efficiency for AR-NECP (MLP) and AR-NECP (NE) under reweighting across different domains distribution and $\alpha$ values. These results come from Mistral7B.}
\resizebox{\linewidth}{!}{
\begin{tabular}{c|c|ccccc}
\hline
\textbf{Distribution} & \textbf{Method} & $\alpha$=0.05 & $\alpha$=0.10 & $\alpha$=0.15 & $\alpha$=0.20 & $\alpha$=0.25 \\
\hline
\multirow{2}{*}{whole} 
& AR-NECP (MLP)  & 4.16 & 2.45 & 1.91 & 1.66 & 1.46 \\
& AR-NECP (NE)   & 3.67 & 2.35 & 1.89 & 1.63 & 1.45 \\
\hline
\multirow{2}{*}{en-d} 
& AR-NECP (MLP)  & 8.06 & 3.05 & 2.14 & 1.71 & 1.47 \\
& AR-NECP (NE)   & 6.95 & 2.78 & 2.09 & 1.70 & 1.47 \\
\hline
\multirow{2}{*}{gg-d} 
& AR-NECP (MLP)  & 2.89 & 2.06 & 1.85 & 1.64 & 1.41 \\
& AR-NECP (NE)   & 2.73 & 2.02 & 1.80 & 1.63 & 1.40 \\
\hline
\end{tabular}}
\label{tab:mre}
\end{table}
\begin{table}[h]
\centering
\caption{Comparison of prediction efficiency for AR-NECP (MLP) and AR-NECP (NE) under reweighting across different domains distribution and $\alpha$ values. These results come from Gemma7B. }
\resizebox{\linewidth}{!}{
\begin{tabular}{c|c|ccccc}
\hline
\textbf{Distribution} & \textbf{Method} & $\alpha$=0.05 & $\alpha$=0.10 & $\alpha$=0.15 & $\alpha$=0.20 & $\alpha$=0.25 \\
\hline
\multirow{2}{*}{whole} 
& AR-NECP (MLP)  & 13.9 & 8.97 & 3.38 & 2.46 & 1.94 \\
& AR-NECP (NE)   & 13.25 & 6.36 & 2.91 & 2.13 & 1.76 \\
\hline
\multirow{2}{*}{en-d} 
& AR-NECP (MLP)  & 16.44 & 11.11 & 5.19 & 2.74 & 2.0 \\
& AR-NECP (NE)   & 19.9 & 12.91 & 8.28 & 3.54 & 2.53 \\
\hline
\multirow{2}{*}{gg-d} 
& AR-NECP (MLP)  & 10.66 & 3.42 & 2.43 & 1.92 & 1.73 \\
& AR-NECP (NE)   & 8.82 & 2.72 & 2.08 & 1.78 & 1.61 \\
\hline
\end{tabular}}
\label{tab:gre}
\end{table}
Tables \ref{tab:gre}, \ref{tab:mre}, \ref{tab:lre} are the results from HotpoQA. Under most settings, NE performs better than MLP.
\begin{table}[h]
\centering
\caption{Comparison of prediction efficiency for AR-NECP (NE on top) vs.\ AR-NECP (MLP) under reweighting across distributions and $\alpha$ for Llama8B (HotpotQA).}
\resizebox{\linewidth}{!}{
\begin{tabular}{c|c|ccccc}
\hline
\textbf{Distribution} & \textbf{Method} & $\alpha{=}0.05$ & $0.10$ & $0.15$ & $0.20$ & $0.25$ \\
\hline
\multirow{2}{*}{sc-d} 
& AR-NECP (NE)  & 9.48 & 4.41 & 2.84 & 2.14 & 1.84 \\
& AR-NECP (MLP) & 9.79 & 5.47 & 3.10 & 2.25 & 1.84 \\
\hline
\multirow{2}{*}{ht-d} 
& AR-NECP (NE)  & 9.31 & 4.28 & 2.77 & 2.15 & 1.84 \\
& AR-NECP (MLP) & 9.14 & 5.04 & 3.06 & 2.20 & 1.73 \\
\hline
\multirow{2}{*}{eg-d} 
& AR-NECP (NE)  & 10.09 & 5.70 & 3.11 & 2.30 & 1.87 \\
& AR-NECP (MLP) & 9.72  & 6.03 & 3.17 & 2.34 & 1.84 \\
\hline
\end{tabular}}
\label{tab:lre}
\end{table}

\begin{table}[h]
\centering
\caption{Comparison of prediction efficiency for AR-NECP (NE on top) vs.\ AR-NECP (MLP) under reweighting across distributions and $\alpha$ for Gemma7B (HotpotQA).}
\resizebox{\linewidth}{!}{
\begin{tabular}{c|c|ccccc}
\hline
\textbf{Distribution} & \textbf{Method} & $\alpha{=}0.05$ & $0.10$ & $0.15$ & $0.20$ & $0.25$ \\
\hline
\multirow{2}{*}{sc-d} 
& AR-NECP (NE)  & 9.00 & 4.58 & 2.86 & 2.12 & 1.75 \\
& AR-NECP (MLP) & 10.85 & 6.19 & 3.23 & 2.19 & 1.72 \\
\hline
\multirow{2}{*}{ht-d} 
& AR-NECP (NE)  & 9.14 & 5.02 & 2.90 & 2.10 & 1.72 \\
& AR-NECP (MLP) & 9.75 & 6.05 & 3.05 & 2.16 & 1.74 \\
\hline
\multirow{2}{*}{eg-d} 
& AR-NECP (NE)  & 8.66 & 4.38 & 2.69 & 1.98 & 1.68 \\
& AR-NECP (MLP) & 9.89 & 5.92 & 3.10 & 2.17 & 1.68 \\
\hline
\end{tabular}}
\label{tab:gre}
\end{table}

\begin{table}[h]
\centering
\caption{Comparison of prediction efficiency for AR-NECP (NE on top) vs.\ AR-NECP (MLP) under reweighting across distributions and $\alpha$ for \textbf{Mistral7B} (HotpotQA).}
\resizebox{\linewidth}{!}{
\begin{tabular}{c|c|ccccc}
\hline
\textbf{Distribution} & \textbf{Method} & $\alpha{=}0.05$ & $0.10$ & $0.15$ & $0.20$ & $0.25$ \\
\hline
\multirow{2}{*}{sc-d} 
& AR-NECP (NE)  & 9.00 & 4.58 & 2.86 & 2.12 & 1.71 \\
& AR-NECP (MLP) & 10.85 & 6.19 & 3.23 & 2.19 & 1.72 \\
\hline
\multirow{2}{*}{ht-d} 
& AR-NECP (NE)  & 9.14 & 5.02 & 2.90 & 2.10 & 1.72 \\
& AR-NECP (MLP) & 9.75 & 6.05 & 3.05 & 2.16 & 1.74 \\
\hline
\multirow{2}{*}{eg-d} 
& AR-NECP (NE)  & 8.66 & 4.38 & 2.69 & 1.98 & 1.68 \\
& AR-NECP (MLP) & 9.89 & 5.92 & 3.10 & 2.17 & 1.68 \\
\hline
\end{tabular}}
\label{tab:mre}
\end{table}
We do not present experiments on RQ3 for MMLU, as the improvement brought by our method is limited. Therefore, it would be meaningless to compare the performance between NE and MLP on this dataset.
\subsection{Additional experiment on RQ4}
Each model in trivaQA shares the same $\hat{n}^k_{train}$ and $n^k_{test}$. Below are for HotpotQA in Table \ref{tab:mean_diff}

\begin{table}[]
\centering
\resizebox{0.5\linewidth}{!}{
\begin{tabular}{c|ccccc}
\hline
\textbf{Shift} & \textbf{sc} & \textbf{ht} & \textbf{eg}  \\
\hline
sa-d   & 0.08 & 0.15 & 0.19 \\
hd-d   & 0.24 & 0.08 & 0.11\\
eg-d  & 0.19 & 0.22 & 0.03 \\
\hline
\end{tabular}
}
\caption{Mean relative error $\Delta^k$ between true and estimated domain counts under different test-time shifts. 
\textit{sc-d}, \textit{ht-d} and \textit{eg-d} denote test sets dominated by the \textit{sc}, \textit{ht} and \textit{eg} domains, respectively. Lower values are better.}
\label{tab:mean_diff}
\end{table}
Below are for MMLU in Table \ref{tab:mmlu_diff}

\begin{table}[]
\centering
\resizebox{0.6\linewidth}{!}{
\begin{tabular}{c|cccc}
\hline
\textbf{Shift} & \textbf{sc} & \textbf{ht} & \textbf{eg} & \textbf{eg} \\
\hline
st-d     & 0.01 & 0.03 & 0.10 & 0.08 \\
hu-d     & 0.05 & 0.01 & 0.09 & 0.13 \\
ss-d     & 0.06 & 0.02 & 0.04 & 0.17 \\
ot-d     & 0.09 & 0.06 & 0.17 & 0.05 \\
whole-d  & 0.02 & 0.03 & 0.07 & 0.06 \\

\hline
\end{tabular}
}

\caption{Mean relative error $\Delta^k$ between true and estimated domain counts under different test-time shifts. 
\textit{hu-d}, \textit{ot-d}, \textit{ss-d} and \textit{st-d} denote test sets dominated by the \textit{hu}, \textit{ot}, \textit{ss} and \textit{st} domains, respectively. Lower values are better.}
\label{tab:mmlu_diff}
\end{table}
\subsection{AI assitant usage}
We used ChatGPT to assist with paper revision and coding.
\end{document}